\documentclass{CVM}

\usepackage{multirow}
\usepackage{longtable}
\usepackage{booktabs, makecell}
\usepackage[implicit=false]{hyperref}
\usepackage{colortbl}
\usepackage{hhline}
\pagecolor{white}

\graphicspath{{figures/},{figures/cvm/}}
\definecolor{mygreen}{RGB}{106, 166, 108}
\newcommand{\mengchen}[1]{\textcolor{black}{#1}}
\newcommand{\shixia}[1]{\textcolor{black}{#1}}
\newcommand{\weikai}[1]{\textcolor{black}{#1}}
\newcommand{\head}[1]{\noindent\textbf{#1}}

\def \etal {{\emph{et al}.\thinspace}}
\def \eg {{\emph{e.g}.\thinspace}}

\def\colorbar#1{%
	\color[RGB]{161, 160, 170}
 \vcenter{\hbox{\rule{0.5\dimexpr#1em}{2.5ex}}}\ \color{black}#1%
}
\CVMsetup{
type      = {Research/Review Article},
doi       = {s41095-0xx-xxxx-x},
title     = {Foundation Models Meet Visualizations: Challenges and Opportunities},
author    = {Weikai Yang$^{1}$, Mengchen Liu$^{2}$, Zheng Wang$^{1}$, and Shixia Liu$^{1}$\cor{}\\},
runauthor = {W. Yang, M. Liu, Z. Wang, and S. Liu},
abstract  = {
Recent studies have indicated that foundation models, such as BERT and GPT, excel in adapting to a variety of downstream tasks.
This adaptability has established them as the dominant force in building artificial intelligence (AI) systems. 
As visualization techniques intersect with these models, a new research paradigm emerges. 
This paper divides these intersections into two main areas: visualizations for foundation models (VIS4FM) and foundation models for visualizations (FM4VIS). 
In VIS4FM, we explore the primary role of visualizations in understanding, refining, and evaluating these intricate models.
This addresses the pressing need for transparency, explainability, fairness, and robustness. 
Conversely, within FM4VIS, we highlight how foundation models can be utilized to advance the visualization field itself.
The confluence of foundation models and visualizations holds great promise, but it also comes with its own set of challenges. 
By highlighting these challenges and the growing opportunities, this paper seeks to provide a starting point for continued exploration in this promising avenue.
},
keywords  = {Visualization; artificial intelligence; machine learning; foundation models; VIS4FM; FM4VIS},
copyright = {The Author(s)},
}

\begin{document}

\maketitle

    \begin{figure}[b] \vskip -2mm
    \small\renewcommand\arraystretch{1.3}
        \begin{tabular}{p{80.5mm}} \toprule\\ \end{tabular}
        \vskip -4.5mm \noindent \setlength{\tabcolsep}{1pt}
        \begin{tabular}{p{3.5mm}p{80mm}}
    $1\quad $ & School of Software, Tsinghua University, Beijing 100084, China. E-mail: \{yangwk21, zheng-wa19\}@mails.tsinghua.edu.cn, shixia@tsinghua.edu.cn.\\
    $2\quad $ & Microsoft, Redmond 98052, United States. E-mail:  mengcliu@microsoft.com.\\
&\hspace{-5mm} Manuscript received: 2022-01-01; accepted: 2022-01-01\vspace{-2mm}
    \end{tabular} \vspace {-3mm}
    \end{figure}

\section{Introduction}\label{sec:introduction}   

\begin{figure}[!htb]
\centering
\includegraphics[width=0.48\textwidth]{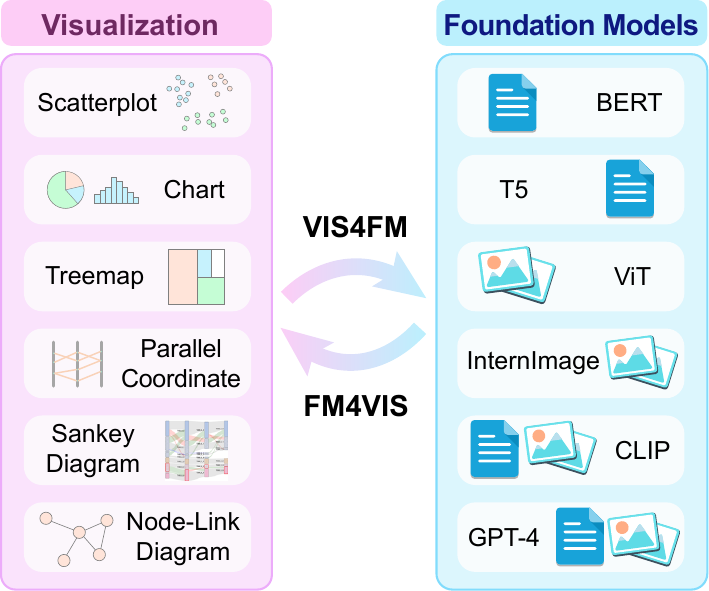}
\caption{The intersections between visualizations and foundation models are divided into two categories: VIS4FM and FM4VIS.}
\label{fig:intersection}
\end{figure}
A foundation model \shixia{is a large-scale machine learning model} that is trained on \shixia{a huge amount of data} across different domains, \weikai{generally} \shixia{using self-supervision}~\cite{bommasani2021opportunities}.
Notable examples of such models include BERT~\cite{devlin2019bert}, InternImage~\cite{wang2022internimage}, CLIP~\cite{Radford2021LearningTV}, and GPT series models~\cite{brown2020language,ouyang2022training,openai2023gpt4}.
They typically possess parameters ranging from hundreds of millions to billions or even trillions.
These immense scales of parameters and training data enable foundation models to capture general knowledge about the world and serve as a ``foundation'' to \shixia{effectively adapt to a variety of downstream tasks, such as natural language understanding, image recognition, question answering, and image segmentation}~\cite{bommasani2021opportunities}.
\shixia{Due to their adaptability, foundation models have become a leading force in shaping the creation of versatile, high-performing AI systems across multiple applications.}
A recent OpenAI report indicates that \shixia{approximately 19\% of jobs have undergone considerable changes, with at least 50\%} of the tasks affected by these models~\cite{eloundou2023gpts}.


In the era of big data and artificial intelligence, there is an increasing need to visualize large-scale datasets and machine learning models for efficient analysis.
\shixia{Recent studies have indicated that incorporating humans into the analysis process can make visualization techniques a critical bridge to human comprehension of complex models}~\cite{liu2017towards,choo2018visual,hohman2018visual, yuan2021survey,Sacha19VIS4ML,wang2021survey,wu2021ai4vis}.
This enhanced human-AI collaboration facilitates effective insight communication, informed decision-making, and improved AI trustworthiness.
When visualization techniques \shixia{meet} foundation models, a new research paradigm emerges.
As shown in Fig.~\ref{fig:intersection}, the intersections spark two promising research areas: visualizations for foundation models (VIS4FM) and foundation models for visualizations (FM4VIS).
In VIS4FM, visualizations become an indispensable mechanism for facilitating the understanding, analysis, and refinement of foundation models. 
Conversely, FM4VIS focuses on how foundation models can be employed to improve visualization techniques by adapting them to different visualization-related tasks, from \shixia{automatically generating} visualizations to communicating richer insights with users.
Embracing these intersections between foundation models and visualizations will \shixia{advance both fields and improve the collaboration between humans and AI}.

While the confluence of foundation models and visualizations holds great potential, it also introduces challenges along with opportunities.
On the one hand, the increasing scale and complexity of foundation models make them difficult to analyze and interpret in traditional manners.
This prompts the need for novel visualization techniques tailored for these large-scale models.
On the other hand, while foundation models have shown the capability to unlock new dimensions of visualization, it is still underexplored how to maximize their capability and seamlessly integrate humans and AI in developing visualizations, which deserves further investigation.
\shixia{This paper highlights both the challenges and opportunities in this emerging research \shixia{topic} and invites further research.
}

\begin{figure*}[!htb]
\centering
\includegraphics[width=\textwidth]{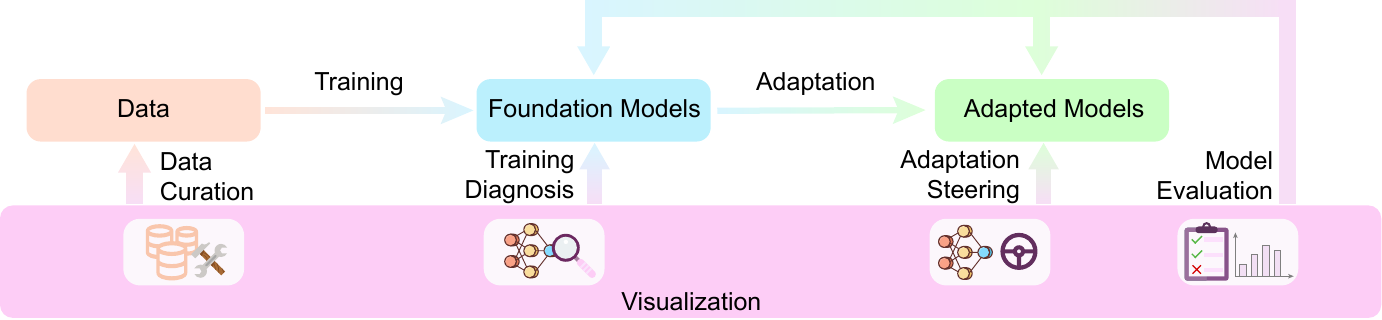}
\caption{How visualizations enhance foundation models along the learning pipeline.}  
\label{fig:vis4fm}
\end{figure*}

\begin{table*}[!htbp]
    \caption{An overview of the four main processes in VIS4FM.}
    \centering
    \renewcommand{\arraystretch}{1.74}
    \begin{tabular}{lllll}
        \toprule
        Process & Tasks Supported by VIS & Description & Examples &\# Examples \\
        \cmidrule(l){1-5}
        & Data Generation & \makecell[l]{use visualizations to help\\ create or augment datasets} & \cite{reif2023visualizing} & $\colorbar{1}$ \\ \cmidrule(l){2-5}
        & Data Integration & \makecell[l]{use visualizations to help\\ integrate data from multiple sources} & - & 0\\ \cmidrule(l){2-5}
        & Data Selection & \makecell[l]{interactively select representative samples \\that align well with the tasks} & - & 0 \\\cmidrule(l){2-5}
        \multirow{-3}{*}{\makecell[l]{Data\\ Curation}} & Data Correction & \makecell[l]{interactively improve \\ the quality of datasets} & \cite{jin2023shortcutlens}\cite{chen2020oodanalyzer}\cite{yang2020diagnosing}\cite{liu2018crowsourcing}\cite{xiang2019interactive}\cite{bauerle2020classifier} & $\colorbar{6}$ \\
        \cmidrule(l){1-5}
        & Model Explanation & \makecell[l]{understand the working\\ mechanism of models} &  \cite{li2021t3}\cite{derose2020attention}\cite{li2023does}\cite{yeh2023attentionviz}  &  $\colorbar{4}$ \\\cmidrule(l){2-5}
        & Performance Diagnosis & \makecell[l]{ troubleshoot issues where models\\ do not perform as expected} &  \cite{li2022unified}\cite{zhang2023sliceteller} & $\colorbar{2}$\\ \cmidrule(l){2-5}
        \multirow{-3}{*}{\makecell[l]{Training\\ Diagnosis}} & Efficiency Diagnosis & \makecell[l]{ identify efficiency bottlenecks\\ in the training process} &  \cite{wei2023visual} & $\colorbar{1}$ \\
        \cmidrule(l){1-5}
        & Model Fine-tuning & \makecell[l]{ analyze what knowledge the\\ models learn during fine-tuning} &  \cite{wang2023commonsensevis}\cite{sevastjanova2022visual} & $\colorbar{2}$\\ \cmidrule(l){2-5}
        & Prompt Engineering & \makecell[l]{facilitate the construction\\ of effective prompts} &  \cite{strobelt2022interactive}\cite{wu2023scattershot}\cite{feng2023promptmagician}\cite{wu2022promptchainer}\cite{wu2022ai} & $\colorbar{5}$ \\ \cmidrule(l){2-5}
        \multirow{-3}{*}{\makecell[l]{Adaptation\\ Steering}} & Alignment via Human Feedback & \makecell[l]{ utilize human feedback \\ to steer model outputs} &  \cite{chung2022talebrush} & $\colorbar{1}$ \\
        \cmidrule(l){1-5}
        & Quantitative Evaluation & \makecell[l]{ use visualizations to present\\ quantitative measures}&  \cite{alsallakh2014visual}\cite{Ren2017Squares}\cite{gortler2022neo} & $\colorbar{3}$ \\\cmidrule(l){2-5}
        \multirow{-2}{*}{\makecell[l]{Model\\ Evaluation}} & Qualitative Evaluation & \makecell[l]{ use visualizations to evaluate and \\ interpret model capability and behaviors} &  \cite{chen2024unified} & $\colorbar{1}$ \\
        \bottomrule
    \end{tabular}
    \label{tab:vis4fm}
\end{table*}

\section{Overview}\label{sec:scope} 
The intersections between visualizations and foundation models contain two aspects: VIS4FM and FM4VIS.

\subsection{VIS4FM}  
VIS4FM focuses on harnessing the power of visualization tools to understand, refine, and evaluate these intricate foundation models.
\weikai{As illustrated in Fig.~\ref{fig:vis4fm}}, foundation models undergo two primary phases: training and adaptation~\cite{bommasani2021opportunities}. 


The data serves as the basis for building foundation models and plays a critical role in determining the performance, reliability, and ethical standing of the resulting models.
Therefore, it is crucial to ensure the data is of high quality, such as \weikai{broad coverage and precise annotation}~\cite{Liu2018Steering,jiang2019recent,chen2021interactive,chen2022towards}.
Given that foundation models often have billions or even trillions of parameters, they have the capacity to learn from vast datasets and absorb both the beneficial and problematic aspects of the data. 
\weikai{Consequently, it is necessary to ensure that the data is not only extensive but also of high quality.
Visualizations facilitate this \textbf{data curation} process from the following four aspects.}
First, visualizations guide the data generation process through real-time feedback on data coverage and correctness.
This allows for immediate adjustments to \weikai{ensure that the generated data adequately represents the intended scope and has correct annotations.}
\weikai{Second, visualizations are useful for integrating heterogeneous data from multiple sources into a coherent and high-quality dataset, which is required to train successful foundation models.}
Third, visualizations assist in data selection by offering a visual representation of the dataset. 
This simplifies the identification of high-quality samples. 
The feedback users provided through visualizations is utilized to further refine the dataset.
Fourth, visualizations disclose anomalies or biases in the data and enable more targeted corrections. 
This improves both the efficiency and accuracy of correcting data.

Training is the initial phase in building foundation models.
During training, the model is trained on vast datasets, which often include diverse and general information.
This allows the model to learn a wide range of features, patterns, and knowledge from the data. 
In this phase, visualizations are essential for \weikai{\textbf{training diagnosis}}~\cite{Liu2017TowardsDeepCNN,Liu2018Analyzing}.
The first task, model explanation, reveals the working mechanism of foundation models.
The second task, performance diagnosis, helps model developers identify the root cause for low performance and make necessary refinements.
The third task, efficiency diagnosis, identifies bottlenecks that impair the training speed or waste resources during training. 

To optimize the performance for specific downstream tasks, the foundation model is usually adapted using task-specific datasets.
This adaptation process refines the model's general knowledge to align more closely with the desired outputs of the tasks. 
In this phase, visualizations are employed to \weikai{facilitate the \textbf{adaptation steering} process} in three ways: model fine-tuning, prompt engineering, and alignment via human feedback.
In model fine-tuning, visualizations help understand \weikai{the knowledge learned by models} and analyze whether it is suitable for the downstream task.
Model developers are then able to compare multiple \weikai{fine-tuned} models and choose the optimal one with a more comprehensive understanding.
In prompt engineering, visualizations streamline the trial-and-error process in crafting effective prompts that lead to the desired outputs.
In alignment via human feedback, the model is steered toward human preferences based on human feedback.
Visualizations serve two functions: they either aid in collecting human feedback to improve training data or offer an interactive platform to iteratively refine the model outputs.


In addition, visualizations serve as a useful technique in enhancing the \weikai{\textbf{model evaluation}} process for both foundation models and adapted models~\cite{chen2024unified}.
For the quantitative evaluation with clear metrics, visualizations offer users a comprehensive and intuitive understanding of model performance.
In addition, given the adaptability of foundation models to a variety of downstream applications, it is also important to evaluate their performance across multiple tasks.
Well-designed visualizations facilitate an efficient comparative analysis based on different metrics, which enables users to select the optimal model or gain insights for further refinement.
For the qualitative evaluation that lacks clear metrics, visualizations serve as a valuable tool for incorporating human judgment into the evaluation process. 
For example, when dealing with open-ended questions that lack definitive ground-truth answers, visualizations can summarize frequent patterns in model-generated answers and provide an informative overview.
This enables users to evaluate \shixia{the quality of these responses} more efficiently.
\shixia{Once low-quality responses are identified, various strategies can be employed to enhance their quality. 
One such method is to enrich the dataset with varied instances of the associated problematic questions.}

\begin{figure*}[!htb]
\centering
\includegraphics[width=\textwidth]{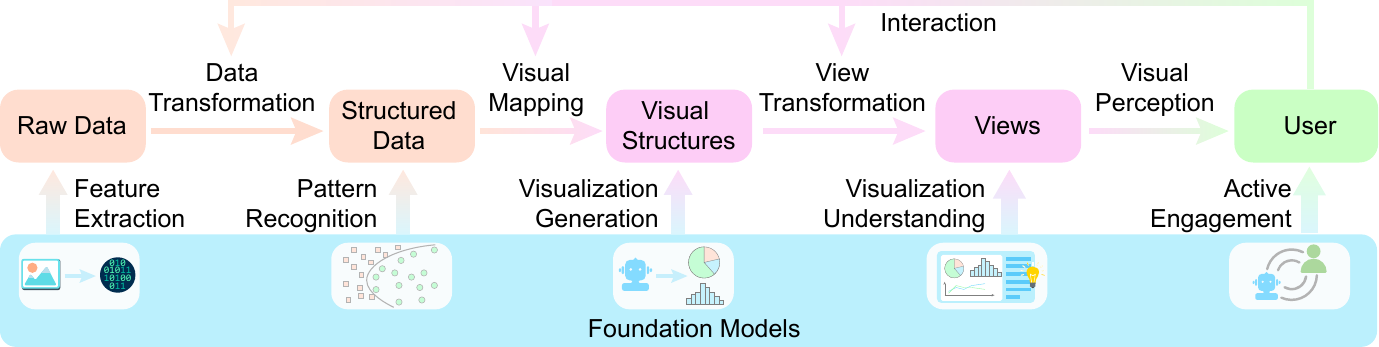}
\caption{How foundation models enhance visualizations along the visualization pipeline.}
\label{fig:fm4vis}
\end{figure*}

\begin{table*}[!htbp]
    \caption{An overview of the four main processes in FM4VIS.}
    \centering
    \renewcommand{\arraystretch}{1.74}
    \begin{tabular}{lllll}
        \toprule
        Process & Tasks Supported by FM & Description & Examples & \# Examples \\
        \cmidrule(l){1-5}
        & Feature Extraction & \makecell[l]{extract informative features \\from unstructured data } & \makecell[l]{\cite{sun2022erato}\cite{ying2022metaglyph}\cite{guo2023edit}\cite{tu2022phrasemap}\\\cite{li2021nbsearch}\cite{narechania2021vitality}\cite{shi2022medchemlens}}& $\colorbar{7}$\\ \cmidrule(l){2-5}
        \multirow{-2}{*}{\makecell[l]{Feature Extraction and \\Pattern Recognition}} & Pattern Recognition &\makecell[l]{automatic identification of \\ patterns in data } & \makecell[l]{\cite{resck2022legalvis}\cite{zhang2020teddy}\cite{wu2023liveretro}\cite{ouyang2023leveraging}\\\cite{tu2023sdrquerier}\cite{chen2023iball}\cite{chen2023sporthesia}\cite{tu2021keywordmap}} & $\colorbar{8}$ \\
        \cmidrule(l){1-5}
        &  Content Generation & \makecell[l]{ generate desired \\  visualization content} & \cite{liu2021advisor}\cite{shen2023data} & $\colorbar{2}$ \\ \cmidrule(l){2-5}
         & Style Generation & 
        \makecell[l]{generate desired styles}
        & \cite{xiao2023let} & $\colorbar{1}$ \\
        \cmidrule(l){2-5}
        \multirow{-3}{*}{\makecell[l]{Visualization\\ Generation}} & Interaction Generation & 
        \makecell[l]{generate desired interactions}
        & - & 0 \\
        \cmidrule(l){1-5}
        & Content Extraction & \makecell[l]{ understand and extract\\content from visualization} & \cite{singh2020stl}\cite{ma2021towards}\cite{song2023vividgraph}\cite{chen2019towards} & $\colorbar{4}$ \\ \cmidrule(l){2-5}
        \multirow{-2}{*}{\makecell[l]{Visualization\\ Understanding}} & Information Communication & \makecell[l]{ summarize and communicate \\ underlying information}& \cite{sultanum2023datatales}\cite{liu2023autotitle}\cite{song2023gvqa} & $\colorbar{3}$ \\
        \cmidrule(l){1-5}
        & Direct Interaction Enhancement & \makecell[l]{ directly enhance user interactions} & \cite{adhikary2021text} & $\colorbar{1}$ \\ \cmidrule(l){2-5}
        \multirow{-2}{*}{\makecell[l]{Active\\ Engagement}} & Predictive Interaction Enhancement & \makecell[l]{ understand user intent to \\ predict the next interaction }&-&0\\
        \bottomrule
    \end{tabular}
    \label{tab:fm4vis}
\end{table*}

\shixia{Building on the above discussion, Table~\ref{tab:vis4fm} offers a summary of the four main processes in VIS4FM.
This table not only provides an outline of existing initiatives but also highlights areas where future research could be beneficial, especially where there has been little to no effort.}


\subsection{FM4VIS}
FM4VIS harnesses the power of foundation models to create more adaptive, user-friendly, and intelligent visualization techniques and systems.
\weikai{These efforts} aim to further advance the visualization field.
\weikai{As illustrated in Fig.~\ref{fig:fm4vis},} the visualization pipeline transforms raw data into an interpretable visual representation that allows users to interact with and derive insights from the presented information~\cite{card1999readings}.
FM4VIS focuses on enhancing each phase in this pipeline: from data transformation and visual mapping to view transformation and visual perception.

Data transformation converts raw data into a format that is more suitable for visualization and analysis.
As foundation models are trained on diverse datasets, they can be used \weikai{to perform \textbf{feature extraction}, which extracts meaningful features from complex data for visualization.}
This is particularly useful for unstructured data, such as text or images, where traditional feature engineering methods \shixia{often produce less informative features}~\cite{zhou2023comprehensive}.
Foundation models can perform tasks like classification, relationship extraction, and object detection to extract various patterns, such as relationships, trends, and anomalies. 
These tasks provide visualization tools with richer pattern data, and thus enable multi-faceted understanding and analysis. 

Visual mapping determines the way to visually represent the underlying data. 
The key is to map the data and their values to marks (\eg, point, line, area) and visual channels (\eg, position, color, size), respectively. 
Foundation models can enrich this phase by facilitating \weikai{\textbf{visualization generation}, including automatic content generation, style generation, and interaction generation.}
These models have the ability to learn patterns and user preferences from datasets, so they can recommend or generate optimal layouts that highlight important data trends. 
Moreover, they can understand the context of the data and suggest appropriate marks and visual channels. 
For example, these models can determine which color palettes best differentiate data categories or decide which shapes represent specific data points more effectively. 
By leveraging foundation models, more insightful and contextually relevant visual representations of data can be generated.
\weikai{With the capability of code generation, foundation models also have the potential to augment visualizations with rich interactions.}

The view transformation converts the abstract visual representation into concrete pixels on a screen.
It is a crucial step to ensure that the final visual representation is effectively communicated to users.
During this phase, foundation models play an important role in \weikai{\textbf{visualization understanding}, which aims to} enhance the understanding of the visualization content \weikai{and communicate the underlying information to users.}
First, they contribute to distilling and abstracting key information from visual presentations. 
For example, a foundation model can be fine-tuned to extract an adaptable visualization template from a set of complex timeline visualizations~\cite{chen2019towards}. 
This involves not only recognizing visual elements, but also understanding their hierarchical and relational significance.
Second, they amplify the user's comprehension of visualizations \weikai{by conveying the key information to users \shixia{in an engaging, multimodal format, such as a combination of natural language with visual elements}.}
For example, these models can provide clear and accurate captions that the visualization designer wants to communicate through the visualizations~\cite{liu2023autotitle}.

Visual perception is a cognitive process that happens in users' minds. 
It interprets the visual representation and translates the colors, shapes, and patterns back into an understanding of the underlying data. 
Moreover, users can interact with the visualizations, such as by zooming, panning, or selecting specific data points, which promotes deeper understanding and reveals further insights from the data.
In this phase, foundation models can \weikai{achieve \textbf{active engagement}} that enhances user interactions from two aspects: direct interaction enhancement and predictive interaction enhancement. 
Direct interaction enhancement employs foundation models to directly simplify user interactions. 
For example, in the context of 3D scatterplots, they can refine the shape of a lasso selection to make them more precise and contextually relevant~\cite{chen2019lassonet}. 
Beyond visual selections, these models can also interpret text descriptions provided by users. 
For instance, when a user describes a specific pattern or attribute, they can process this description and highlight the corresponding visual patterns on the display.
Predictive interaction enhancement uses foundation models to predict and enhance user interactions for immediate responses and broader data exploration insights. 
The predictive capabilities of these models can be harnessed to predict user actions within visualizations.
For example, after observing user interactions with a scatter plot, they can predict where the user is likely to click next to streamline their exploration process~\cite{ottley2019follow}.
Furthermore, a more advanced utilization of these models is to analyze user interactions. 
By observing how a user interacts with visualizations, they can predict not only their imminent actions but also broader attributes, such as their likely performance on a specific task or even \shixia{specific} aspects of their personality~\cite{brown2014finding}.

\shixia{Based on the aforementioned discussion, an overview of the four main processes in FM4VIS is summarized in Table~\ref{tab:fm4vis}. 
In addition to providing an overview of existing efforts, this table also indicates potential future research directions where little or no effort is made}.

\section{Existing VIS4FM Efforts}\label{sec:vis4lfm}

In this section, we introduce recent efforts in VIS4FM, with a focus on data curation, training diagnosis, adaptation steering, and model evaluation (Fig.~\ref{fig:vis4fm}).
\shixia{Typical examples in each category are presented in Table~\ref{tab:vis4fm}.}

\subsection{Data Curation}

Visualization can \shixia{simplify} the data curation process in four \shixia{aspects}: \shixia{data generation, data integration, data selection, and data correction}.
\weikai{Existing efforts mainly focus on data generation and data correction.}

\head{Data Generation}.
\shixia{Data generation is a process of creating new data based on existing data by using large-scale machine learning models that have been trained on a huge amount of data from different domains. 
It plays a crucial role in improving machine learning datasets by employing techniques such as filling in missing values, balancing class distributions, and augmenting sparse data collections.
}
With their capabilities for content generation, foundation models boost efficiency and effectiveness in generating datasets that \shixia{can be used to} train, fine-tune, or test models.
However, these automatically generated datasets usually contain some quality issues, such as the presence of undesirable repetitions and incorrect information (\eg, incorrect annotation, out-of-range value, spurious relationship).
Undesirable repetitions refer to samples that are either highly similar or identical to the seed samples used in dataset generation.
These repetitions may hinder the diversity of the generated dataset.
To address this issue, Reif~\etal\cite{reif2023visualizing} developed LinguisticLens, a visualization tool to identify potential undesirable repetitions in the generated dataset.
This tool organizes similar sentences into clusters based on their syntactic and lexical information.
The clustering results allow users to analyze the linguistic patterns and individual sentences of each cluster more efficiently.
Based on a comprehensive understanding, they can determine whether these similar sentences are valuable enhancements or undesirable repetitions.


\head{Data Correction}.
Data correction refers to the process of correcting noisy annotation and spurious correlations between inputs and outputs (shortcuts) within training datasets.
In the context of traditional deep learning models, many visual analysis methods have been developed to improve both the effectiveness and efficiency of such data correction processes, including improving instance representativeness~\cite{chen2020oodanalyzer,yang2020diagnosing,jin2023shortcutlens} and enhancing annotation quality~\cite{liu2018crowsourcing,xiang2019interactive,bauerle2020classifier}.
Given their emphasis on data-centric issues, these methods are readily transferable for enhancing data quality during the adaptation of foundation models.
For example, ShortCutLens~\cite{jin2023shortcutlens} facilitates the identification of shortcuts in natural language understanding datasets.
This tool provides an overview of potential shortcuts and allows users to analyze samples associated with specific shortcuts.
Once identified, these shortcuts can be addressed by constructing new samples, modifying existing ones, or removing those that are misleading.
Another exemplary work is DataDebugger developed by Xiang~\etal\cite{xiang2019interactive}, which employs a hierarchical visualization to \shixia{facilitate} the examination and correction of annotations.
By using \shixia{this visualization}, users can navigate through the dataset, identify the samples of interest, and provide accurate annotations on them.
These annotations are then propagated to correct other noisy annotations using an annotation correction algorithm, \shixia{thereby reducing human efforts}.


\subsection{Training Diagnosis}

Based on \shixia{the main analytical focus}, existing VIS4FM efforts in training diagnosis can be \shixia{divided} into three categories: model explanation, performance diagnosis, and efficiency diagnosis.\looseness=-1

\head{Model Explanation}.
Model explanation refers to the process of interpreting the working mechanism of machine learning models and how they make decisions.
In recent years, transformer-based foundation models, such as BERT and Vision Transformer, have achieved remarkable performance across various tasks~\cite{li2021t3,li2023does,derose2020attention,yeh2023attentionviz}.
While the success of these models is often attributed to self-attention, its working mechanism remains somewhat unclear.
To bridge this gap, DeRose~\etal\cite{derose2020attention} developed Attention Flows to interpret how attention flows across tokens and contributes to the final prediction results.
In addition, it supports the comparison of attention flows between two models to enable the analysis of their similarities and differences.
Li~\etal\cite{li2023does} proposed a visual analysis tool tailored for analyzing Vision Transformer models.
This tool offers a multi-faceted examination of attention, including the importance of different attention heads, the attention strengths across different image patches, and the attention patterns that individual heads learn.
While \shixia{these} methods \shixia{are effective in} interpreting working mechanisms based on individual samples, analyzing the patterns across multiple samples provides a more comprehensive perspective.
To this end, Yeh~\etal\cite{yeh2023attentionviz} introduced AttentionViz, a tool designed to examine the self-attention patterns across multiple input samples simultaneously.
It first projects the query and key vectors used by transformer models into a shared space.
By examining these query-key interactions in the shared space, model developers can \shixia{better} understand the behavior of different attention heads.

\head{Performance Diagnosis}.
Performance diagnosis aims to troubleshoot issues where models do not perform as expected and understand the reason behind them.
Compared with the model explanation, it focuses more on \mengchen{diagnosing} performance issues rather than \mengchen{explaining} model working mechanisms.
Visualization techniques provide an interactive and intuitive environment to streamline the performance diagnosis process.
For example, Li~\etal\cite{li2022unified} developed DeepNLPVis to identify and diagnose performance issues in deep natural language processing models.
DeepNLPVis introduces an information-based sample interpretation method to extract the intra-word and inter-word information.
The corpus-level, sentence-level, and word-level visualizations are tightly integrated to visually explain model behavior.
With a comprehensive understanding of how the model processes inputs, model developers can identify and address performance issues efficiently.
SliceTeller~\cite{zhang2023sliceteller} allows model developers to diagnose model performance on different subsets of validation data.
It first automatically constructs several subsets of data with potential performance issues and presents them for performance diagnosis.
After model developers identify critical subsets for further optimization, SliceTeller estimates the performance changes across different subsets.
\shixia{This enables} developers to compare the trade-offs and decide whether to accept the optimization.

\head{Efficiency Diagnosis}.
In contrast to performance diagnosis, efficiency diagnosis focuses on identifying efficiency bottlenecks that slow down the training speed or consume unnecessary resources during the training process.
As foundation models continue to grow in scale, the importance of the efficiency diagnosis becomes even more critical.
A widely used strategy to accelerate the \shixia{training} of a foundation model is to parallelize \shixia{the process} in a distributed cluster.
Despite its effectiveness, it is challenging to diagnose the parallel training process due to the intricate nature of parallelization strategies and the large volume of profiling data\shixia{, such as execution time, resource utilization, and communication overhead.}
To tackle these issues, Wei~\etal\cite{wei2023visual} proposed a visual analysis method for diagnosing parallel training processes.
This method integrates detailed information about the parallelization strategy into the computational graph, which is visualized using a directed acyclic graph layout.
To facilitate the analysis of the profiling data, they developed an enhanced Marey’s graph to visualize the execution time of operators, the peak memory of different devices, and inter-device communication latency.
Additionally, an aggregation method is employed to handle the large volume of profiling data within the Marey’s graph.


\subsection{Adaptation Steering}
Based on the methods to align models with human preferences, existing VIS4FM efforts in adaptation steering can be divided into three categories: model fine-tuning, prompt engineering, and alignment via human feedback.

\head{Model Fine-tuning}.
Model fine-tuning is a widely used technique for adapting foundation models to downstream tasks by updating model parameters using task-specific training data.
In model fine-tuning, model developers want to understand what knowledge the models learn and whether this knowledge is suitable for the downstream tasks.
Visualizations have been shown effective in providing insights into model behavior~\cite{Liu2017TowardsDeepCNN,Wexler2020WhatIf,wang2023commonsensevis}\shixia{, so they serve as a useful method to accelerate the fine-tuning process}.
For example, Wang~\etal\cite{wang2023commonsensevis} developed CommonsenseVIS to analyze what commonsense knowledge the models learn and whether the knowledge is used in model reasoning.
It first employs a knowledge graph to extract commonsense knowledge from input data.
\shixia{The alignment of model behavior with human reasoning is then achieved by utilizing the overlap between the extracted knowledge and the knowledge learned by the model.}
\shixia{Through the use of interactive visualizations for the alignment, model developers can} understand and diagnose issues where the models fall short in learning effectively.
Beyond fine-tuning entire foundation models, there is a growing trend toward parameter-efficient methods, such as adapter~\cite{houlsby2019parameter} and low-rank adaptation (LoRA)~\cite{hu2021lora}.
These methods \shixia{add a few task-specific parameters to the foundation models and train only those new parameters}.
By doing so, it not only reduces training complexity but also allows adapters and LoRA modules to learn task-specific knowledge without modifying the weights of foundation models.
As a result, there are many publicly available adapters and LoRA modules fine-tuned on different tasks and datasets~\cite{adapterhub}.
Understanding what task-specific knowledge is learned can facilitate model developers in selecting an appropriate adapter or LoRA module for their tasks.
For example, Sevastjanova~\etal\cite{sevastjanova2022visual} proposed a visual analysis method to support the comparison of knowledge learned by different adapters.
It integrates three types of explanation methods: concept embedding similarity, concept embedding projection, and concept prediction similarity, and uses them to compare different adapters.
This method enables developers to make informed decisions about which adapter best suits the downstream task of interest.

\head{Prompt Engineering}.
Instead of traditional fine-tuning methods, foundation models can also be adapted to downstream tasks using prompting techniques.
A prompt is a natural language description of the task, which makes the task suitable for foundation models to handle.
However, the choice of prompts can significantly influence model performance, and designing a high-performing prompt requires deep expertise.
To alleviate the burden of manually crafting prompts, Strobelt~\etal\cite{strobelt2022interactive} developed PromptIDE that facilitates users in constructing different prompts, comparing their performance, and interactively refining them.
By specifying the range of variables in a prompt template, a comprehensive set of prompts is generated that spans all potential combinations.
These generated prompts are then evaluated on a small set of validation data with ground-truth labels to provide quantitative measures.
Users can then compare their performance and refine the prompt template or a single prompt.
In a similar vein, ScatterShot~\cite{wu2023scattershot} focuses on helping users interactively select informative samples and add them to the prompt.
It employs a clustering technique to organize samples into clusters based on task-specific key phrases and offers performance estimation for each cluster.
Low-performance clusters are prioritized for further exploration and sample selection.
For tasks without clear quantitative measures, such as text-to-image generation, visualizations can assist in exploring the relationships between input prompts and output results.
For example, PromptMagician~\cite{feng2023promptmagician} streamlines the interactive refinement of text prompts in text-to-image generation tasks.
It employs a prompt recommendation model to retrieve prompt-image pairs that are similar to the input prompt from a pre-existing database.
The retrieved pairs are visualized in a 2D space using t-SNE and organized using hierarchical clustering for efficient exploration.
Furthermore, important and relevant prompt keywords are also extracted to facilitate the prompt refinement process.
Recently, the chain-of-thought technique has also emerged as an effective strategy for enhancing the performance of foundation models in handling complex tasks~\cite{wei2022chain}.
A chain of thought is a series of prompts that breaks down a complex task into a sequence of more manageable sub-tasks.
Visual analysis tools can aid users with limited experience in authoring their own chains~\cite{wu2022promptchainer,wu2022ai}.
For example, Wu~\etal\cite{wu2022ai} developed AIChains, which supports eight primitive operations \shixia{well-suited for language models}.
An interactive interface is designed to \shixia{facilitate users in examining and analyzing} the chain structure and the model outputs.
Based on the analysis, they can make adjustments in different granularities, ranging from refinement within an individual prompt to modifying the intermediate model outputs and even restructuring the entire chain.

\head{Alignment via Human Feedback}.
In contrast to model fine-tuning and prompt engineering, model alignment directly utilizes human feedback to steer model outputs toward human preferences.
Visualization techniques are suitable for collecting human feedback and communicating the associated changes in model output.
Through this human-in-the-loop process, users can iteratively align the model outputs with their preferences.
Recently, Talebrush~\cite{chung2022talebrush} is developed to support writers in iteratively crafting stories. 
It employs line-sketching interactions alongside a GPT-based language model to support writers \shixia{in dictating character fortune in line with their creative goals}.
\shixia{Writers may refine the generated narrative by editing the text or modifying their initial sketch.}\looseness=-1

\subsection{Model Evaluation}
Foundation models can be evaluated in two ways: quantitative evaluation and qualitative evaluation. 

\head{Quantitative Evaluation}.
Quantitative evaluation employs pre-defined quantitative measures to evaluate model performance.
Various visualization techniques have been developed to enrich the presentation of these quantitative measures, thereby offering a comprehensive and intuitive understanding of model performance~\cite{alsallakh2014visual,Ren2017Squares,gortler2022neo}.
For example, G{\"o}rtler~\etal\cite{gortler2022neo} developed Neo, which extends traditional confusion matrices to facilitate the evaluation of classification tasks with complex label structures.
Users can efficiently explore confusion matrices related to hierarchical or multi-output labels and then inspect model confusion patterns.

\head{Qualitative Evaluation}.
Qualitative evaluation lacks clear metrics and \shixia{often relies on visualizations to integrate} human judgment into the evaluation process.
For example, Chen~\etal\cite{chen2024unified} developed a unified evaluation method suitable for a variety of tasks in computer vision, including image classification, object detection, and instance segmentation.
In addition to revealing class-level confusion patterns, it also facilitates a fine-grained examination of model capability and behaviors at the sample level.
For example, \shixia{when users} visually compare the model-generated segmentation masks with ground-truth masks, \shixia{they frequently observe inadequate segmentation of helicopter rotors.
This observation guides them to enhance model performance by incorporating a boundary-based loss specifically for helicopter segmentation}.\looseness=-1


\section{Existing FM4VIS Efforts}\label{sec:LFM4VIS}

In this section, we introduce recent efforts in FM4VIS, with a focus on feature extraction and pattern recognition, visualization generation, visualization understanding, and active engagement (Fig.~\ref{fig:fm4vis}).
\shixia{Typical examples in each category are presented in Table~\ref{tab:fm4vis}.}
\looseness=-1

\subsection{Feature Extraction and Pattern Recognition}

\head{Feature Extraction}.
Feature extraction transforms unstructured data, such as text and images, into semantic feature vectors. 
\shixia{Foundation models, pre-trained on vast datasets, often outperform traditional models in this task~\cite{bommasani2021opportunities}}. 
These high-quality semantic feature vectors facilitate advanced visualization techniques. 
Methods for enhancing visualization include querying relevant data~\cite{sun2022erato,ying2022metaglyph,narechania2021vitality,tu2022phrasemap,li2021nbsearch,guo2023edit} and \shixia{enriching with metadata}~\cite{shi2022medchemlens}.
For example, Erato~\cite{sun2022erato} is a human-machine cooperative system for generating data stories.
Once users decide on several key data facts of the story that they want to focus on, Erato utilizes an interpolation algorithm to generate intermediate data facts that smoothly connect different key data facts.
To achieve this, it fine-tunes a BERT model to generate high-quality fact embedding for fact interpolation.
Similarly, MetaGlyph~\cite{ying2022metaglyph} utilizes a pre-trained sentence-BERT to transform both the descriptions of data attributes and data topics into semantic features.
MetaGlyph then calculates the distances between these features and ranks the attributes according to the distances between the attribute descriptions and the data topic.
Those attributes with smaller distances are prioritized to be selected and subsequently visualized.

\head{Pattern Recognition}.
\shixia{Pattern recognition utilizes extracted features to identify a range of patterns that enhance both understanding and analysis}.
Similar to existing methods that employ traditional machine learning models, foundation models are also utilized to perform various tasks, such as classification~\cite{resck2022legalvis,zhang2020teddy,wu2023liveretro,ouyang2023leveraging,tu2023sdrquerier}, 
object detection~\cite{chen2023iball,chen2023sporthesia}, and relationship extraction~\cite{tu2021keywordmap}.
For example, LegalVis~\cite{resck2022legalvis} employs a fine-tuned Longformer model to identify binding precedents (past legal decisions made by higher courts) in legal documents.
Similarly, Teddy~\cite{zhang2020teddy} utilizes a fine-tuned BERT model to extract fine-grained opinions (\eg, cleanliness, service) from the review text and convey them to data scientists.

\subsection{Visualization Generation}
Foundation models have been utilized to facilitate the visualization generation process by either directly generating visualization content (\eg, visualization type, data encoding, annotations)~\cite{liu2021advisor,shen2023data} or generating visualization styles (\eg, color scheme, layout style, typography)~\cite{xiao2023let}.

\head{Content Generation}.
Content generation uses foundation models \shixia{to produce} desired visualization content.
For example, Liu~\etal\cite{liu2021advisor} developed ADVISor to generate visualizations with annotations given tabular data and natural language questions.
In ADVISor, a BERT model is first fine-tuned to extract the features of both questions and table heads.
Building upon these features, several lightweight models are trained to decide the selected attributes, aggregation types, visualization types, and annotations that best address the provided questions.
The corresponding visualization is generated based on this information.
Data Player~\cite{shen2023data} is another representative work designed to simplify the creation of data videos based on the input static visualizations and the corresponding narration text.
It uses OpenAI gpt-3.5-turbo to establish semantic connections between visualization components and narrative entities.
These semantic connections are then utilized to generate narration-animation interplay in the resulting data videos.\looseness=-1

\head{Style Generation}.
Foundation models have also been leveraged to produce the desired visualization style.
Xiao~\etal\cite{xiao2023let} developed ChartSpark to simplify the generation of chart visualizations in pictorial style.
It employs a text-to-image diffusion model to generate the corresponding visualization style given the semantic text prompts.
In addition, it can also take a chart image as an additional input to ensure that the generated visualization is in close alignment with the given chart.
To further enhance the quality of the final outputs, users can utilize image-to-image generation techniques to improve the harmony and consistency of the generated charts.

\subsection{Visual\shixia{ization} Understanding}
Existing efforts can be classified into two categories: content extraction and information communication.

\head{Content Extraction}.
Content extraction focuses on extracting important content from visualizations, including data content~\cite{ma2021towards,singh2020stl,song2023vividgraph} and visualization templates~\cite{chen2019towards}.
In the vein of extracting data content from visualizations,
Ma~\etal\cite{ma2021towards} fine-tuned several models to classify chart types, analyze legends, and detect different visual elements such as boxes and points.
These detected elements are converted back into data values based on the legend information.
For extracting visualization templates, Chen~\etal\cite{chen2019towards} utilized deep learning models to segment and extract visual elements from timeline infographics.
The extracted graphical elements are used as visualization templates for creating similar infographics with different data.

\head{Information Communication}.
With the capability in content generation, foundation models serve as valuable tools for communicating the extracted content and underlying information to users~\cite{song2023gvqa,sultanum2023datatales,liu2023autotitle}.
For example, Sultanum~\etal\cite{sultanum2023datatales} proposed DataTales to create data-driven articles based on data visualizations.
It takes charts as inputs and leverages OpenAI gpt-3.5-turbo to generate corresponding narratives and titles.
These generated narratives are then linked back to the original chart to improve readability and the overall comprehension of the data being presented.
Liu~\etal\cite{liu2023autotitle} developed AutoTitle, an interactive tool designed to interactively generate meaningful titles for visualizations.
It first extracts underlying data from the visualizations and then computes high-level facts through operations like aggregation and comparison.
Based on the computed facts, a foundation model T5~\cite{raffel2020T5} is fine-tuned to generate fluent and informative natural language descriptions.

\subsection{Active Engagement}
Foundation models offer a promising \shixia{way} to understand user intent and refine their interaction results.
For example, in virtual environments, entering text without input devices is challenging and usually contains many errors.
By leveraging a BERT model to re-rank possible word alternatives in the user's text input, the word error rate is significantly reduced~\cite{adhikary2021text}.
In addition to refining the interaction results, some efforts aim to simplify the interaction process, for example, by employing natural language~\cite{wang2022towards}.


\section{Research Opportunities}
In this section, we explore potential avenues for research in both VIS4FM and FM4VIS. 
Specifically, we focus on identifying underexplored potential and new challenges to offer a straightforward roadmap for future studies.
\subsection{VIS4FM}

\subsubsection{Data Curation}

\head{Data Generation}.
Foundation models have shown the capability to generate training datasets for specific tasks.
The automatically generated datasets may contain several quality issues, including undesirable repetition, low coverage, and incorrect annotations.
Although there is an initial effort to address undesirable repetition~\cite{reif2023visualizing}, the issues of low coverage and incorrect annotations are still underexplored.
For the issue of low coverage, visualizations \shixia{offer a useful way to} explore the distribution of the generated datasets and identify the regions with insufficient training samples.
Based on the findings, users can interactively steer the data generation strategy to generate more samples in those regions.
For the issue of incorrect annotations, visualizations \shixia{serve as a powerful tool for users to enhance data quality. 
For example, with appropriate visualizations, they can easily identify specific subsets where} the samples tend to contain noisy annotations.
These corrections then act as valuable feedback for foundation models and contribute to the generation of more accurate data.
In addition, the issue of incorrect annotations can also be addressed by using data selection, \weikai{which is also facilitated by visualizations and} introduced below.

\head{Data Integration}.
Foundation model training usually requires the collection \weikai{and preprocessing} of vast amounts of data from multiple sources.
Merging these heterogeneous data into a coherent and high-quality dataset poses considerable complexities, such as handling data inconsistencies and resolving semantic differences across different sources.
These issues often see improvement through human feedback during the integration process.
In this context, visualization techniques usually play a crucial role in facilitating a more efficient data integration and governance process.
One interesting avenue for future research is to develop a visualization-guided preprocessing framework that enables interactive adjustments to the preprocessing procedure and continuous monitoring of data integrity.
Another promising avenue lies in the investigation of novel visualization techniques that simultaneously handle both the large-scale and heterogeneous nature of training data.
These techniques would make it easier to compare data distributions from different sources and identify inconsistencies.

\head{Data Selection}.
The training and adaptation of foundation models are computationally intense and usually require millions or \shixia{even billions} of training data~\cite{openai2023gpt4}.
This large-scale data requirement \weikai{introduces complexities} in several aspects, including data storage, computational power, and processing time.
Furthermore, training of foundation models is becoming a serious source of carbon emissions threatening our environment~\cite{schwartz2020green}.
Recent studies have shown that selecting a subset of data \weikai{for training} can achieve comparable or even better performance~\cite{raffel2020T5,zhou2023lima}.
These findings suggest the possibility of reducing the computational and environmental costs associated with model training.
Visualizations serve as valuable tools for exploring large-scale datasets and selecting high-quality training data~\cite{zhou2024cluster,Yang2020Interactive}.
However, there are two major challenges that need to be addressed.

The first challenge revolves around scalability, which is particularly significant in the context of foundation models. 
The huge amount of data for training and fine-tuning these models is too big to fit in memory and makes it difficult to process and visualize all the data at once.
This not only calls for out-of-memory sampling techniques but also poses real-time interaction challenges for visualization. 
One possible solution is to initially present an overview of the data distribution using these out-of-memory sampling techniques.
This method allows users to quickly examine the general landscape and identify regions that warrant closer inspection. 
Users can then zoom into these targeted regions for a more granular analysis. 
As the new data is not loaded from memory, it is worth studying how to support real-time interactions.

The second challenge stems from the unannotated and unstructured nature of the training data.
Most training data for foundation models are unstructured data without annotation, such as images or text crawled from websites.
Their unannotated nature makes it difficult to evaluate the quality of training data and select high-quality samples for training.
One possible solution is to design multiple metrics to visually summarize the data characteristics from different perspectives. 
Their unstructured nature poses difficulties for users in quickly understanding the content of samples, which requires innovative visualizations of the data to alleviate the cognitive load.
In addition, multimodal data is now widely used in training foundation models.
However, visualizations of alignments between different modalities remain underexplored and deserve further investigation.

The selection of test data shares most of the challenges as the selection of training data, such as scalability issues and the unstructured nature of the data.
However, there are some differences worth noting.
The primary goal of test data is to faithfully reflect the performance of foundation models while also exposing their potential weaknesses.
Therefore, it is essential that the test data cover both the common samples that models process regularly and the ``edge case'' samples where models may fail.
Visualization techniques are suitable for \shixia{examining the selection balance between these two types of samples.
It is therefore worth exploring how to integrate visualization techniques with the subset selection method for a well-balanced selection}.

\subsubsection{Training Diagnosis}

\head{Model Explanation}.
The intrinsic nature of foundation models is defined by their vast number of parameters. 
This vastness, while being the source of their capability, also makes them challenging to interpret. 
It is daunting to comprehend the myriad interactions, transformations, and computations that these parameters undergo. 
When a foundation model produces outputs, it is the result of a cascade of intricate operations influenced by millions or even billions of parameters. 
Tracing back through these operations to identify the exact reasoning or mechanism is similar to navigating a vast, complex maze without a map. 
The larger and more complex the model, the harder it becomes to interpret the specific factors or processes that led to the given output.

The aforementioned challenge posed by the scale and complexity of foundation models demands innovative visualization solutions to incorporate human knowledge into the analysis process. 
There is a growing opportunity to design and develop novel visualization techniques tailored for such large-scale models. 
These visualization tools can serve as ``lenses,'' which allow users to peer into the depths of these models and offer insights that can be grasped intuitively.
Additionally, exploration based on rich interaction techniques is also important for foundation model explanations. 
These exploration methods would aim to simplify, without losing the essence, the complex behaviors of foundation models into more understandable forms. 
The goal is a delicate balance between the accessibility and faithfulness of the explanation. 
This might involve developing multi-level interpretation mechanisms, where users can choose the granularity of the explanation, or harnessing unsupervised techniques to automatically identify and present the most salient features or operations driving the model's decisions.
Multi-level interpretation mechanisms are designed to offer explanations at varying levels of detail, from a high-level overview to detailed, granular insights. 
At the highest level, these explanations provide a general summary of the decision-making logic of the model. 
This is referred to as \textbf{surface-level interpretation}.
For example, for a text generation task, a surface-level explanation might state, ``The model generated this sentence based on the overall sentiment of the input.''
It can also provide a summary of the associated statistics, such as confidence and bias scores.
The next level can provide \textbf{component-level interpretation}, which aims to explain the role of specific model components, such as particular layers or attention heads. 
For example, ``The 10th attention head focused primarily on the relationship between subject and object in the sentence.''
The potential deepest level may provide \textbf{parameter-level interpretation.} 
It enables the examination of the influence and interactions of specific parameters or groups of parameters. 
This could involve visualizing weights, gradients, or activations associated with particular tokens or features.
Given the vast amount of data present at each level, there is a pressing need for an effective sampling method that can easily capture human interest and display corresponding data.
This motivates the study of interactive sampling strategies, which requires the development of interactive visualizations to facilitate the detection of different user intents.\looseness=-1

\head{Online Training Diagnosis}.
With the increasing complexity of foundation models, the training time of foundation models usually takes weeks or even months on high-end GPUs.
Tradition offline methods gather relevant data after the training process and then feed them into the analysis tool, which is \shixia{less effective in reducing unnecessary training trials}.
Moving the visual analysis earlier in the model development workflow can save vast amounts of time and computing resources, such as halting ineffective and inefficient training immediately.
Therefore, it is necessary to develop visualization techniques that are suitable for monitoring real-time running results and identifying performance issues and/or efficiency issues.
There are two interesting avenues that deserve exploration.

The first promising avenue is to support an in-depth analysis of model performance during model training.
While some existing efforts like Tensorboard~\cite{abadi2016tensorflow} have supported the online monitoring of the training process, they only considered high-level performance metrics, such as loss and prediction accuracy.
These metrics are too abstract to effectively troubleshoot the reason why the model does not perform as expected.
To tackle this issue, it is necessary to integrate advanced data analysis and model analysis modules into visualizations to provide richer information.
By analyzing the sample content and how the model processes them, model developers can gain more insights into the performance issues and address them accordingly.

The second promising avenue lies in the management of large-scale profiling data in online diagnosis.
Given the rapid generation of profiling data and the input/output overhead associated with transferring data from GPU to memory \weikai{or even} disk storage, it becomes impractical to store all the data and then transfer them to the visualization tool for analysis.
In-situ visualization is one of the promising methods to address this issue~\cite{ma2009situ}.
It generates visualizations directly within the computational environment where the data is generated.
Although in situ visualization has been shown to be useful in scientific visualization~\cite{rapp2022image,richer2022scalability}, it is still underexplored whether it can be employed to streamline the efficiency diagnosis during the model training.

\subsubsection{Adaptation Steering}

\head{Model Fine-tuning}.
After fine-tuning a foundation model for a specific task, the model will deviate from its pre-trained version.
The changes are not only about performance metrics but also include model behavior, such as how the model processes different types of input or forms new input-output associations.
By analyzing these behavior changes, model developers can understand how generic knowledge evolves into task-specific knowledge and identify where the model does not work as expected.
Therefore, a promising research opportunity lies in using visualizations to effectively monitor these behavior changes and disclose abnormal behavior during the fine-tuning process.
With an in-depth understanding of the behavior changes, model developers can identify when the model starts to exhibit biases or vulnerabilities that downgrade the model performance.
After identifying these negative issues, visualizations offer an efficient way to interactively steer the fine-tuning process, \shixia{for example, by adding more balanced or targeted data}.
This method enhances not only the model performance but also the reliability and robustness.\looseness=-1



\head{Prompt Engineering}.
Recent studies have shown that providing some high-quality examples within the prompts can greatly enhance model performance, which is known as the in-context learning ability~\cite{dong2022survey}.
In-context learning is a valuable component of prompt engineering. 
In this setup, prompt engineering becomes a critical exercise in curating and structuring examples that can guide the model effectively. 
To fully leverage the capabilities of foundation models and achieve satisfying performance, the provided examples should be well-suited for the downstream task.
However, generating high-quality examples requires expertise and often involves iterative refinement, which is usually trial-and-error in nature.
Visualizations offer an efficient way to facilitate this refinement process by \shixia{introducing humans into the analysis loop}~\cite{Liu2018Visual,choo2018visual,yuan2021survey}.
One promising solution involves employing visualizations to visually illustrate model responses across different in-context examples.
The insights derived from the visualizations enable users to evaluate the effectiveness of the constructed examples and identify those most suitable for the current task.
Based on the findings, they can make informed refinements to the examples for better performance.
In addition to interactively refining examples for each task, another promising direction lies in using visualizations to summarize general principles for example selection~\cite{yuan2022visual}.
By exploring different subsets of examples and performing comparative analysis among them, users can summarize the principles of which types of examples are beneficial and which are not.
These principles contribute to a more systematic and informed example selection to craft effective prompts for the downstream task.

\head{Alignment via Human Feedback}.
In the model adaptation process, aligning the model behavior with human preferences is essential.
This alignment not only improves the user experience by generating more relevant responses, but also tackles ethical and societal concerns~\cite{ouyang2022training}.
Recently, reinforcement learning from human feedback has been shown to be effective in aligning model behavior with human preferences~\cite{ouyang2022training}.
This method first trains a reward model directly from human feedback, which predicts whether the response aligns with human preferences (high reward) or not (low reward).
Subsequently, this reward information guides the optimization of the foundation models through reinforcement learning.
In this process, the key lies in collecting high-quality human feedback and utilizing this data to train a reward model that accurately captures human preferences.
Visualization techniques are suitable for both tasks.
First, 
interactive visualizations have already demonstrated their value in enhancing the process of collecting human feedback. 
Existing research on interactive data labeling showcases the effectiveness of visualization techniques in facilitating the collection of human-generated data~\cite{khayat2019vassl,bernard2018towards,yang2022diagnosing}.
Second, visualization techniques offer an efficient way to diagnose the training process of reward models and interactively refine them through additional human feedback.
By tightly integrating human inputs into this process, reward models are better aligned with actual human preferences, thereby providing more accurate and reliable reward information for the ongoing optimization of the foundation model.\looseness=-1

The challenge in this context is multi-faceted. 
First, collecting high-quality human feedback is a difficult task in itself, and the difficulty is amplified when the data must be fed into a reward model that drives reinforcement learning. 
Errors or biases in feedback collection can result in skewed training and unreliable models. 
Second, while visualization techniques offer the opportunity to collect human-generated data more effectively, integrating these techniques seamlessly with reinforcement learning pipelines presents its own set of complexities. 
Balancing real-time interaction with computational efficiency in an already complex training process is a challenge.
\looseness=-1

\head{Model Selection}.
\weikai{In recent years, more and more model developers have uploaded their fine-tuned models together with meta-data (\eg, descriptions, model architectures, resources requirements) to a learnware market~\cite{zhou2022learnware,adapterhub,huggingface}.}
The growing availability of publicly fine-tuned foundation models opens a new door to the efficient development of AI systems.
When confronted with an AI task, users have the option of searching and selecting a pre-existing model \weikai{in the learnware market} that fits their specific needs.
However, without adequate expertise, they might find it difficult to navigate through the large model space to identify the optimal foundation model~\cite{Wang2020Visual}.
The primary challenge is to guide user exploration by capturing user needs and recommending models that have the potential to achieve high performance.
Accordingly, a possible solution is to employ visualization techniques to visually illustrate the model space.
Through these visualizations, users can navigate the complex model space more easily, understand model behaviors, identify model limitations, and compare models from multiple perspectives, such as performance scores and resources requirements.
This multi-faceted understanding and comparison enable them to identify the optimal model for their specific tasks.

\subsubsection{Model Evaluation}
\shixia{The visualization field has extensively covered quantitative evaluation. 
Therefore, we focus on discussing the research challenges and opportunities related to qualitative evaluation.}

\head{Evaluating Free-form Outputs}.
Recently, foundation models have achieved impressive performance in various tasks, notably in answering open-ended questions without definitive ground-truth answers.
However, evaluating the quality of free-form model responses remains challenging due to the high variability in possible responses and the absence of clear ground-truth answers.
\shixia{Addressing this challenge requires human involvement in the evaluation process.
However, the sheer volume of data makes it unfeasible for users to manually inspect and assess each model response individually.}
One possible solution is to semi-automatically create rules for evaluating the model responses, which can be achieved by active learning methods.
Visualizations enhance this process by offering a comprehensive overview of these evaluation rules and their associated model responses.
Users can then iteratively refine these rules according to their preferences, which ultimately leads to more accurate and reliable evaluations.
Another potential solution is \shixia{to utilize visualizations} to highlight the responses that are challenging for the semi-automatic evaluation methods and present them to users for manual review.
To minimize redundancy and simplify this process, it is essential to cluster a massive volume of responses and intuitively summarize the clustering results in visualizations.
\looseness=-1

\head{Robustness}.
Many foundation models, such as those in the GPT series~\cite{brown2020language,openai2023gpt4}, are generative models.
Although these models demonstrate impressive comprehension or generation abilities, they may also misinterpret inputs or generate off-target or wrong outputs. 
Such inconsistencies pose challenges in the reliable deployment of these models, especially in scenarios where a single error could have significant consequences.
As a result, there is an urgent need to get a clear picture of their robustness.
With this information, \shixia{users can assess how well these models might perform in different situations and target weak areas for fine-tuning}~\cite{Cao2020Analyzing,liu2018robustness}.

To achieve this, one possible solution is to construct a set of input samples with perturbations and compare the corresponding model responses with well-designed visualizations.
This helps users understand how small changes in the input can affect the model output.
This provides information on its robustness and sensitivity. 
Visualizations can assist in identifying critical inputs that deserve closer examination, interactively constructing perturbated inputs, and summarizing multiple model responses for efficient analysis.
Another solution is to analyze a large number of inputs collected in real-world scenarios to identify potential robustness issues among them.
In many applications, models are deployed in complex environments where they encounter a wide range of inputs.
Manually examining each for robustness issues can be an overwhelming task.
Visualizations offer an effective means to explore and filter a set of similar inputs that produce diverse outputs.
These anomalies \shixia{often signal potential issues with robustness}. 
Once these \shixia{anomaly pairs} are identified, visualization tools help run ``what-if'' analyses. 
These analyses examine how the model behaves under various conditions, thereby identifying specific areas where the model's robustness can be improved.









\head{Fairness}.
Given that foundation models are increasingly deployed in diverse cultural contexts and used by diverse user groups, it is crucial to prioritize culturally sensitive, ethically sound, and socially aligned explanations provided by VIS4FM techniques. 
Consequently, it becomes essential to explore how VIS4FM techniques effectively navigate cross-cultural differences, address ethical dilemmas, and assess their broader societal impact. 
This research direction is essential to advance the area of VIS4FM and ensure responsible model deployments.\looseness=-1 

First, cross-cultural differences can significantly impact how individuals perceive and interpret information. 
Cultural factors such as language, beliefs, values, and norms influence the understanding and acceptance of foundation models and their explanations. 
Therefore, it is important to investigate how VIS4FM techniques can account for and adapt to cross-cultural differences in explanation generation and presentation. 
This involves studying cultural biases in foundation models, developing culture-aware explanation methods, and conducting user studies in diverse cultural contexts to assess the effectiveness and appropriateness of VIS4FM techniques.

Second, ethical considerations are important in the development and application of adapted models. 
Visualization techniques should adhere to ethical principles such as transparency, fairness, privacy, and accountability.
This includes addressing issues such as algorithmic bias, discrimination, and the potential impact of VIS4FM explanations on vulnerable populations.
Researching specific ethical frameworks and guidelines for VIS4FM can help ensure that the deployment of adapted models with visual explanations is done in a responsible and ethical way.

\subsection{FM4VIS}


\subsubsection{Feature Extraction and Pattern Recognition} 
Foundation models offer two notable new opportunities compared with traditional machine learning models in these two processes.
\shixia{First,} due to their training on more diverse and extensive datasets, foundation models \shixia{typically generate} features of higher quality than those obtained from traditional machine learning models.
These high-quality features better disclose the underlying patterns in the data, such as clusters~\cite{Radford2021LearningTV,zhang2020teddy,qiu2022docflow}.
\shixia{These features facilitate} visualization researchers in designing suitable visualizations \shixia{to} analyze the data.
Second, 
previous feature extraction \shixia{methods} mainly focus on single-modality data, such as Latent Dirichlet Allocation (LDA) for textual data~\cite{blei2003latent} and scale-invariant feature transform (SIFT) for image data~\cite{lowe1999object}.
There are some recent research efforts in training multi-modality foundation models, such as CLIP~\cite{Radford2021LearningTV}, to map multi-modality data into one \shixia{unified} feature space.
This enables visualization researchers to \shixia{design} a unified visualization for multi-modality data,
\shixia{which} facilitates in disclosing the inter-modality relationships within the data.

\subsubsection{Visualization Generation}

\head{Prompted Content Generation}.
Large language models, as widely studied foundation models, have shown the capabilities of generating source code given natural language prompts.
For example, Code LLAMA~\cite{roziere2023code} has shown state-of-the-art performance on several public code generation benchmarks, such as MBPP~\cite{austin2021program}.
\shixia{An interesting avenue for future research could be to democratize visualization design by extending these capabilities to automatically generate advanced visualizations.
By integrating with well-known engines, such as D3~\cite{bostock2011d3} and matplotlib~\cite{hunter2007matplotlib}, this method simplifies the process for individuals without prior experience in visualization design. 
They can create their own advanced visual data representations and address complex data challenges.
Although the execution of this concept seems intuitive using existing public APIs, the concept is not fully implemented}.
\shixia{Several research efforts are still underway to improve the quality of the generated visualizations.} 
First, the development of a visualization-related instruction tuning dataset is critical.
Currently, the visualization code, such as the D3 code, only makes up a small portion of the training corpus of large language models.
Thus, developing a dataset containing both instructions and accompanying visualization code is necessary to increase performance in creating different visualization components with large language models. 
\mengchen{The importance of visualization-specific datasets has been demonstrated by existing automatic graph layout methods~\cite{kwon2019deep}.}
Using such datasets and leveraging advanced fine-tuning techniques, such as Reinforcement Learning from Human Feedback, can significantly enhance the model's code-generation capabilities in the visualization field.
\shixia{Second,} prompt engineering is essential to ensure that the generated visualizations align with the user intent. 
Existing research has illustrated that different prompts have a substantial influence over the outputs generated by large language models~\cite{zamfirescu2023johnny}. 
Thus, crafting an effective prompt is critical.
To alleviate human efforts in the tedious prompt curation process, recent techniques, such as automatic prompt optimization~\cite{pryzant2023automatic} can be leveraged.

\head{Style Generation}. 
In computer vision, \shixia{style transfer refers to the technique of applying the visual style of one image to the content of another~\cite{jing2019neural}.
This often involves two images: a content image and a style image. The algorithm reconfigures the content image so it takes on the artistic style found in the style image.} 
For instance, StyleGAN~\cite{abdal2019image2stylegan} leverages generative adversarial networks to distill the style cues from the reference image. 
\shixia{By incorporating style-based generator layers, it offers fine-grained control over image attributes, which improves the quality and versatility of generated images.}
Currently, these style transfer models remain within the domain of natural image generation.
\shixia{However, the principles behind style transfer offer potential applications beyond visual arts. 
They open avenues into other fields like visualization.}
It is still an open but important research avenue to harness style transfer techniques effectively in the visualization field.
Such an \shixia{extension would allow users to easily transfer stylistic elements from one visualization to another.} 
Moreover, it would serve as a valuable resource for those with limited \weikai{programming} skills and would facilitate the creation of user-centric visualizations with minimal effort. 
\shixia{This can make complex data more accessible and understandable to a broader audience}.
A critical challenge in this endeavor is to preserve the data integrity in the transferred visualization.
Unlike natural images, visualization is a visual form of data.
Therefore, the faithful representation of these data is critical.
Current style transfer techniques, when applied to visualization, may introduce subtle changes in visual elements, such as line length adjustments.
This potentially leads to the risk of perceptual errors.
\shixia{A promising research opportunity lies in adapting the style transfer models to incorporate the original data used for generating the visualizations, thereby ensuring data integrity while transferring styles.
Another challenge lies in the automatic recommendation of styles, a task complicated by the multifaceted intricacies of human perception and divergent individual preferences.
For example, one user might prioritize clarity and simplicity, while another might focus on intricate detail and vibrant color schemes. 
Additionally, cultural background, professional training, and even mood can influence what a user finds engaging or easy to interpret. 
These varying factors make the automatic process of recommending styles a complex endeavor, as the system must account for a wide range of subjective preferences.
}


\head{Interaction Generation}.
Interaction enables users to tailor views to specific information needs.
It serves as a cornerstone for effective data exploration and analysis. 
However, creating intuitive and responsive interactions is a challenge that demands expertise in both visualization techniques and programming. 
The code-generation capabilities of foundation models offer a significant opportunity.
One interesting avenue for research is the simplified interaction design.
Similar to the prompted content generation we discussed before, users can implement some basic interactions by describing their intent using natural languages.
The challenge here lies in the ambiguity that natural language often presents.
This makes it difficult to clearly describe complex interactive functionalities.
Therefore, an exciting opportunity exists in extending foundation models to accept other types of input, such as sketches or video examples, to produce more accurate interaction designs.
On a more advanced level, foundation models have the potential to simplify the programming of complex interactions, such as multi-stage animation scheduling and sophisticated visual effects. 
Here, ensuring that the generated code meets quality standards remains an ongoing issue.
In response to this, a potential avenue for future research is the development of automatic quality assurance mechanisms that can evaluate and refine the code generated by foundation models.\looseness=-1


\subsubsection{Visualization Understanding}
\head{Content Extraction}.
Previous research has highlighted the enhanced reasoning capabilities inherent in foundation models~\cite{brown2020language}.
\shixia{Using these capabilities}, visualization researchers can then adapt foundation models to comprehend complex visualizations, such as node-link diagrams or treemaps, \shixia{and extract key information for in-depth analysis}.
For example, when presented with a node-link diagram representing \weikai{a complex social network, foundation models can effectively identify key information such as influential users, sub-communities, and their connections.}
Descriptive captions and concise summaries of \weikai{this information can be generated and presented alongside the visualization, which greatly facilitates visualization comprehension.}
A critical challenge in adapting current foundation models to understand complex visualizations is the lack of domain-specific data.
Currently, existing public datasets in the visualization field often focus on simple charts like bar charts or line charts~\cite{wang2021survey}.
Thus, it is critical to create a public dataset containing complex visualizations and their extracted insights.
\weikai{Another challenge lies in identifying 
\shixia{contextually relevant information that matches} the analytical focus.
Interactive visualizations often excel at conveying useful patterns embedded in a large amount of data.
For example, a visualization of a social network may present multiple interesting sub-communities that deserve exploration.
A tailored summary of sub-communities of interest is often more beneficial than a generic overview of the entire network. 
Consequently, the task of capturing users' analytical focus and dynamically extracting relevant patterns and tailored summaries for visualizations emerges as a promising avenue for future investigation.
}

\head{Visual-Question-Answering-Based Communication}.
In computer vision, developing machine learning models to answer questions about an image is an active research topic, which is referred to as visual question answering~\cite{antol2015vqa}.
With the help of foundation models, it becomes possible for users to engage in free-form and open-ended dialogues about visualizations, which \weikai{alleviates the cognitive load of understanding visualizations}.
To achieve this vision, two key aspects deserve consideration.
First, the model needs to have a robust linguistic comprehension capability and possess a large amount of knowledge to effectively address open-ended questions about visualization.
While some foundation models like PaLM2 have achieved remarkable accuracy rates exceeding 90$\%$ on the CommonsenseQA benchmark dataset~\cite{anil2023palm}, \weikai{the capability of answering open-ended questions about visualization remains a topic for further study.}
Second, contextual awareness is a critical component to enable a smooth multi-round dialog experience in foundation models.
Currently, chat-centric models, such as ChatGPT, have demonstrated the ability to deliver desired results conditioned on previous user prompts \weikai{in the dialog}~\cite{ouyang2022training}.
Adding the underlying data to the prompts can help the foundation model understand the visualizations more precisely and answer numerical questions.
\weikai{However, incorporating data into the prompts raises scalability issues.}
Directly incorporating all the data into the prompts is not only inefficient, but may also be unfeasible \shixia{given the large volume of the data}.
To solve this problem, the development of data abstraction techniques (\eg, sampling~\cite{zhao2021preserving,yuan2020evaluation}, statistical summary) becomes necessary, which enables the extraction of the most important data closely linked to the generated visualizations.



\subsubsection{Active Engagement}

\head{Direct Interaction Enhancement}.
Currently, several widely used interactions, such as brushing and zooming, have been overlooked in the training of foundation models.
Consequently, these models struggle to understand and enhance such user interactions.
To bridge this gap, there are two potential solutions.
\shixia{A straightforward solution} is to convert these interactions into formats that current foundation models can readily understand. 
\weikai{For example, mouse-click interactions can be converted into textual descriptions and fed into large language models.}
A more promising solution involves training or adapting foundation models to understand these interactions directly. 
Encouragingly, initial efforts have emerged to enhance model capabilities in this direction.
For example, DragGAN enables users to manipulate objects within images through drag-and-drop interactions~\cite{pan2023drag}.
Such efforts are notable steps toward expanding the capabilities of interaction-aware foundation models.\looseness=-1



\head{Predictive Interaction Enhancement}.
Recently, there are several initiatives to enhance the capabilities of foundation models by creating foundation-model-based AI agents~\cite{wang2023survey}.
These AI agents are designed to \shixia{mimic} human behaviors and typically include various modules, such as perception, memory, planning, and reflection, each of which is often supported by a foundation model.
Such agents can actively identify human feedback and incorporate it into their reflection module, which adapts their actions in subsequent steps based on this feedback~\cite{park2023generative}.
It becomes feasible to employ these AI agents for visual analysis tasks.
Traditional approaches require domain experts to manually examine data through visualizations and identify patterns through sequences of interactions, which is time-consuming and expertise-dependent.
In contrast, AI agents possess the potential to simplify this analysis process by generating similar interaction sequences based on the interaction sequences performed by domain experts.
Nevertheless, achieving such productive collaboration between humans and AI agents still poses two challenges.

The first challenge lies in fine-tuning a foundation model that is capable of automatically generating interaction sequences for extracting useful patterns. 
To alleviate the effort to interact with different visual analysis tools,
foundation models can be utilized to generate interaction sequences, which are then used to automatically extract pattern candidates.
Domain experts only need to examine these candidates and find the most relevant patterns for further analysis.
The second challenge lies in efficiently adapting the foundation model to specific visual analysis tools and domain experts.
To achieve this, boosting the model's capacity for in-context learning is crucial. 
The foundation model should be able to learn from a few example interaction sequences performed by experts and then extract more patterns in similar interactions.

\section{Conclusions}\label{sec:conclusions}
The intersections of foundation models and visualizations signify a substantial step in the advancement of AI systems.
On the one hand, VIS4FM becomes crucial in explaining the complexities of foundation models.
This highlights the growing need for transparency, explainability, fairness, and robustness in the expanding role of AI.
On the other hand, FM4VIS provides new pathways to further advance visualization techniques.
While the integration of these two \shixia{fields} presents certain challenges, the potential benefits and advancements they can bring are undeniable.
As we stand at this crossroads, it is essential to confront the challenges head-on while embracing the vast opportunities that lie ahead.
This confluence not only promises a brighter future for AI and visualization, but also encourages a sustained journey of discovery and innovation in this emerging research topic.



\appendix



\subsection*{Declaration of competing interest}

The authors have no competing interests to declare that are relevant to the content of this article.

\subsection*{Fundings}

This work was supported by the National Natural Science Foundation of China under grants (No.s U21A20469, 61936002), the National Key R\&D Program of China under Grant 2020YFB2104100, grants from the Institute Guo Qiang, THUIBCS, and BLBCI.

\subsection*{Authors' contributions}

\noindent Weikai Yang: Conceptualization, Writing - Original Draft, Writing - Review Editing.

\noindent Mengchen Liu: Conceptualization, Writing - Original Draft, Writing - Review Editing.

\noindent Wang Zheng: Writing - Original Draft, Writing - Review Editing.

\noindent Shixia Liu: Conceptualization, Supervision, Writing
– Original Draft, Writing - Review Editing, Funding acquisition.

\subsection*{Acknowledgements}

The authors would like to thank Dr. Xiting Wang, Dr. Changjian Chen, Jun Yuan, Yukai Guo, Jiangning Zhu, and Duan Li for their valuable comments.




\bibliographystyle{CVM}

{\normalsize  \bibliography{ref}}

\begin{thebibliography}{100}
\expandafter\ifx\csname urlstyle\endcsname\relax
  \providecommand{\doi}[1]{doi:\discretionary{}{}{}#1}\else
  \providecommand{\doi}{doi:\discretionary{}{}{}\begingroup
  \urlstyle{rm}\Url}\fi

\bibitem{bommasani2021opportunities}
Bommasani R, Hudson DA, Adeli E, Altman R, Arora S, von Arx S, Bernstein MS,
  Bohg J, Bosselut A, Brunskill E, et~al.. On the opportunities and risks of
  foundation models. \emph{arXiv preprint arXiv:2108.07258}, 2021.

\bibitem{devlin2019bert}
Devlin J, Chang MW, Lee K, Toutanova K.
\newblock {BERT}: Pre-training of deep bidirectional transformers for language
  understanding.
\newblock In \emph{Proceedings of the Conference of the North {A}merican
  Chapter of the Association for Computational Linguistics: Human Language
  Technologies, Volume 1 (Long and Short Papers)}, 4171--4186, 2019.

\bibitem{wang2022internimage}
Wang W, Dai J, Chen Z, Huang Z, Li Z, Zhu X, Hu X, Lu T, Lu L, Li H, et~al..
  InternImage: Exploring large-Scale vision foundation models with deformable
  convolutions. \emph{arXiv preprint arXiv:2211.05778}, 2022.

\bibitem{Radford2021LearningTV}
Radford A, Kim JW, Hallacy C, Ramesh A, Goh G, Agarwal S, Sastry G, Askell A,
  Mishkin P, Clark J, Krueger G, Sutskever I.
\newblock Learning transferable visual models from natural language
  supervision.
\newblock In \emph{International Conference on Machine Learning}, 8748--8763,
  2021.

\bibitem{brown2020language}
Brown TB, Mann B, Ryder N, Subbiah M, Kaplan J, Dhariwal P, Neelakantan A,
  Shyam P, Sastry G, Askell A, Agarwal S, Herbert-Voss A, Krueger G, Henighan
  T, Child R, Ramesh A, Ziegler DM, Wu J, Winter C, Hesse C, Chen M, Sigler E,
  Litwin M, Gray S, Chess B, Clark J, Berner C, McCandlish S, Radford A,
  Sutskever I, Amodei D.
\newblock Language models are few-shot learners.
\newblock In \emph{Advances in Neural Information Processing Systems},
  1877--1901, 2020.

\bibitem{ouyang2022training}
Ouyang L, Wu J, Jiang X, Almeida D, Wainwright CL, Mishkin P, Zhang C, Agarwal
  S, Slama K, Ray A, Schulman J, Hilton J, Kelton F, Miller L, Simens M, Askell
  A, Welinder P, Christiano P, Leike J, Lowe R. Training language models to
  follow instructions with human feedback. \emph{arXiv preprint
  arXiv:2203.02155}, 2022.

\bibitem{openai2023gpt4}
OpenAI. GPT-4 Technical Report. \emph{arXiv preprint arXiv:2303.08774}, 2023.

\bibitem{eloundou2023gpts}
Eloundou T, Manning S, Mishkin P, Rock D. {GPTs} are {GPTs}: An early look at
  the labor market impact potential of large language models. \emph{arXiv
  preprint arXiv:2303.10130}, 2023.

\bibitem{liu2017towards}
Liu S, Wang X, Liu M, Zhu J. Towards better analysis of machine learning
  models: A visual analytics perspective. \emph{Visual Informatics}, 2017,
  1(1): 48--56.

\bibitem{choo2018visual}
Choo J, Liu S. Visual analytics for explainable deep learning. \emph{IEEE
  Computer Graphics and Applications}, 2018, 38(4): 84--92.

\bibitem{hohman2018visual}
Hohman F, Kahng M, Pienta R, Chau DH. Visual analytics in deep learning: An
  interrogative survey for the next frontiers. \emph{{IEEE} Transactions on
  Visualization and Computer Graphics}, 2019, 25(8): 2674--2693.

\bibitem{yuan2021survey}
Yuan J, Chen C, Yang W, Liu M, Xia J, Liu S. A survey of visual analytics
  techniques for machine learning. \emph{Computational Visual Media}, 2021,
  7(1): 3--36.

\bibitem{Sacha19VIS4ML}
{Sacha} D, {Kraus} M, {Keim} DA, {Chen} M. {VIS4ML}: An ontology for visual
  analytics assisted machine learning. \emph{{IEEE} Transactions on
  Visualization and Computer Graphics}, 2019, 25(1): 385--395.

\bibitem{wang2021survey}
Wang Q, Chen Z, Wang Y, Qu H. A survey on {ML4VIS}: Applying machine learning
  advances to data visualization. \emph{{IEEE} Transactions on Visualization
  and Computer Graphics}, 2022, 28(12): 5134--5153.

\bibitem{wu2021ai4vis}
Wu A, Wang Y, Shu X, Moritz D, Cui W, Zhang H, Zhang D, Qu H. {AI4VIS}: Survey
  on artificial intelligence approaches for data visualization. \emph{{IEEE}
  Transactions on Visualization and Computer Graphics}, 2022, 28(12):
  5049--5070.

\bibitem{reif2023visualizing}
Reif E, Kahng M, Petridis S. Visualizing Linguistic Diversity of Text Datasets
  Synthesized by Large Language Models. \emph{arXiv preprint arXiv:2305.11364},
  2023.

\bibitem{jin2023shortcutlens}
Jin Z, Wang X, Cheng F, Sun C, Liu Q, Qu H. ShortcutLens: A visual analytics
  approach for exploring shortcuts in natural language understanding dataset.
  \emph{{IEEE} Transactions on Visualization and Computer Graphics}, 2024,
  30(1): 1--11, accepted by VIS 2023.

\bibitem{chen2020oodanalyzer}
Chen C, Yuan J, Lu Y, Liu Y, Su H, Yuan S, Liu S. {OoDAnalyzer}: Interactive
  analysis of out-of-distribution samples. \emph{{IEEE} Transactions on
  Visualization and Computer Graphics}, 2021, 27(7): 3335--3349.

\bibitem{yang2020diagnosing}
Yang W, Li Z, Liu M, Lu Y, Cao K, Maciejewski R, Liu S.
\newblock Diagnosing concept drift with visual analytics.
\newblock In \emph{{IEEE} Conference on Visual Analytics Science and
  Technology}, 12--23, 2020.

\bibitem{liu2018crowsourcing}
Liu S, Chen C, Lu Y, Ouyang F, Wang B. An interactive method to improve
  crowdsourced annotations. \emph{{IEEE} Transactions on Visualization and
  Computer Graphics}, 2019, 25(1): 235--245.

\bibitem{xiang2019interactive}
Xiang S, Ye X, Xia J, Wu J, Chen Y, Liu S.
\newblock Interactive correction of mislabeled training data.
\newblock In \emph{Proceedings of the IEEE Conference on Visual Analytics
  Science and Technology}, 57--68, 2019.

\bibitem{bauerle2020classifier}
B{\"a}uerle A, Neumann H, Ropinski T. Classifier-Guided Visual Correction of
  Noisy Labels for Image Classification Tasks. \emph{Computer Graphics Forum},
  2020, 39(3): 195--205.

\bibitem{li2021t3}
Li R, Xiao W, Wang L, Jang H, Carenini G.
\newblock T3-Vis: visual analytic for Training and fine-Tuning Transformers in
  NLP.
\newblock In \emph{Proceedings of the Conference on Empirical Methods in
  Natural Language Processing: System Demonstrations}, 220--230, 2021.

\bibitem{derose2020attention}
DeRose JF, Wang J, Berger M. Attention flows: Analyzing and comparing attention
  mechanisms in language models. \emph{{IEEE} Transactions on Visualization and
  Computer Graphics}, 2021, 27(2): 1160--1170.

\bibitem{li2023does}
Li Y, Wang J, Dai X, Wang L, Yeh CCM, Zheng Y, Zhang W, Ma KL. How Does
  Attention Work in Vision Transformers? A Visual Analytics Attempt.
  \emph{{IEEE} Transactions on Visualization and Computer Graphics}, 2023,
  29(6): 2888--2900.

\bibitem{yeh2023attentionviz}
Yeh C, Chen Y, Wu A, Chen C, Vi{\'e}gas F, Wattenberg M. {AttentionViz}: A
  Global View of Transformer Attention. \emph{arXiv preprint arXiv:2305.03210},
  2023.

\bibitem{li2022unified}
Li Z, Wang X, Yang W, Wu J, Zhang Z, Liu Z, Sun M, Zhang H, Liu S. A unified
  understanding of deep nlp models for text classification. \emph{{IEEE}
  Transactions on Visualization and Computer Graphics}, 2022, 28(12):
  4980--4994.

\bibitem{zhang2023sliceteller}
Zhang X, Ono JP, Song H, Gou L, Ma KL, Ren L. {SliceTeller}: A data
  slice-driven approach for machine learning model validation. \emph{{IEEE}
  Transactions on Visualization and Computer Graphics}, 2023, 29(1): 842--852.

\bibitem{wei2023visual}
Wei Y, Wang Z, Wang Z, Dai Y, Ou G, Gao H, Yang H, Wang Y, Cao CC, Weng L,
  et~al.. Visual Diagnostics of Parallel Performance in Training Large-Scale
  DNN Models. \emph{{IEEE} Transactions on Visualization and Computer
  Graphics}, 2024, 30(1): 1--11, accepted by VIS 2023.

\bibitem{wang2023commonsensevis}
Wang X, Huang R, Jin Z, Fang T, Qu H. {CommonsenseVIS}: Visualizing and
  Understanding Commonsense Reasoning Capabilities of Natural Language Models.
  \emph{arXiv preprint arXiv:2307.12382}, 2023.

\bibitem{sevastjanova2022visual}
Sevastjanova R, Cakmak E, Ravfogel S, Cotterell R, El-Assady M. Visual
  comparison of language model adaptation. \emph{{IEEE} Transactions on
  Visualization and Computer Graphics}, 2023, 29(1): 1178--1188.

\bibitem{strobelt2022interactive}
Strobelt H, Webson A, Sanh V, Hoover B, Beyer J, Pfister H, Rush AM.
  Interactive and visual prompt engineering for ad-hoc task adaptation with
  large language models. \emph{{IEEE} Transactions on Visualization and
  Computer Graphics}, 2023, 29(1): 1146--1156.

\bibitem{wu2023scattershot}
Wu S, Shen H, Weld DS, Heer J, Ribeiro MT.
\newblock ScatterShot: Interactive In-context Example Curation for Text
  Transformation.
\newblock In \emph{Proceedings of the International Conference on Intelligent
  User Interfaces}, 353--367, 2023.

\bibitem{feng2023promptmagician}
Feng Y, Wang X, Wong KK, Wang S, Lu Y, Zhu M, Wang B, Chen W. {PromptMagician}:
  Interactive Prompt Engineering for Text-to-Image Creation. \emph{arXiv
  preprint arXiv:2307.09036}, 2023.

\bibitem{wu2022promptchainer}
Wu T, Jiang E, Donsbach A, Gray J, Molina A, Terry M, Cai CJ.
\newblock {Promptchainer}: Chaining large language model prompts through visual
  programming.
\newblock In \emph{Proceedings of the CHI conference on human factors in
  computing systems}, 1--10, 2022.

\bibitem{wu2022ai}
Wu T, Terry M, Cai CJ.
\newblock {AI Chains}: Transparent and controllable human-ai interaction by
  chaining large language model prompts.
\newblock In \emph{Proceedings of the CHI conference on human factors in
  computing systems}, 1--22, 2022.

\bibitem{chung2022talebrush}
Chung JJY, Kim W, Yoo KM, Lee H, Adar E, Chang M.
\newblock TaleBrush: Sketching stories with generative pretrained language
  models.
\newblock In \emph{Proceedings of the CHI conference on human factors in
  computing systems}, 1--19, 2022.

\bibitem{alsallakh2014visual}
Alsallakh B, Hanbury A, Hauser H, Miksch S, Rauber A. Visual Methods for
  Analyzing Probabilistic Classification Data. \emph{{IEEE} Transactions on
  Visualization and Computer Graphics}, 2014, 20(12): 1703--1712.

\bibitem{Ren2017Squares}
Ren D, Amershi S, Lee B, Suh J, Williams JD. Squares: Supporting Interactive
  Performance Analysis for Multiclass Classifiers. \emph{{IEEE} Transactions on
  Visualization and Computer Graphics}, 2017, 23(1): 61--70.

\bibitem{gortler2022neo}
G{\"o}rtler J, Hohman F, Moritz D, Wongsuphasawat K, Ren D, Nair R, Kirchner M,
  Patel K.
\newblock Neo: Generalizing confusion matrix visualization to hierarchical and
  multi-output labels.
\newblock In \emph{Proceedings of the CHI Conference on Human Factors in
  Computing Systems}, 1--13, 2022.

\bibitem{chen2024unified}
Chen C, Guo Y, Tian F, Liu S, Yang W, Wang Z, Wu J, Su H, Pfister H, Liu S. A
  Unified Interactive Model Evaluation for Classification, Object Detection,
  and Instance Segmentation in Computer Vision. \emph{{IEEE} Transactions on
  Visualization and Computer Graphics}, 2024, 30(1): 1--11, accepted by VIS
  2023.

\bibitem{Liu2018Steering}
Liu S, Andrienko G, Wu Y, Cao N, Jiang L, Shi C, Wang YS, Hong S. Steering data
  quality with visual analytics: The complexity challenge. \emph{Visual
  Informatics}, 2018, 2(4): 191--197, \doi{10.1016/j.visinf.2018.12.001}.

\bibitem{jiang2019recent}
Jiang L, Liu S, Chen C. Recent research advances on interactive machine
  learning. \emph{Journal of Visualization}, 2019, 22: 401--417.

\bibitem{chen2021interactive}
Chen C, Wang Z, Wu J, Wang X, Guo LZ, Li YF, Liu S. Interactive Graph
  Construction for Graph-Based Semi-Supervised Learning. \emph{IEEE
  Transactions on Visualization and Computer Graphics}, 2021, 27(9):
  3701--3716.

\bibitem{chen2022towards}
Chen C, Wu J, Wang X, Xiang S, Zhang SH, Tang Q, Liu S. Towards Better Caption
  Supervision for Object Detection. \emph{IEEE Transactions on Visualization
  and Computer Graphics}, 2022, 28(4): 1941--1954.

\bibitem{Liu2017TowardsDeepCNN}
Liu M, Shi J, Li Z, Li C, Zhu J, Liu S. Towards better analysis of deep
  convolutional neural networks. \emph{{IEEE} Transactions on Visualization and
  Computer Graphics}, 2017, 23(1): 91--100.

\bibitem{Liu2018Analyzing}
Liu M, Shi J, Cao K, Zhu J, Liu S. Analyzing the Training Processes of Deep
  Generative Models. \emph{{IEEE} Transactions on Visualization and Computer
  Graphics}, 2018, 24(1): 77--87, \doi{10.1109/tvcg.2017.2744938}.

\bibitem{sun2022erato}
Sun M, Cai L, Cui W, Wu Y, Shi Y, Cao N. {Erato}: Cooperative data story
  editing via fact interpolation. \emph{{IEEE} Transactions on Visualization
  and Computer Graphics}, 2023, 29(1): 983--993.

\bibitem{ying2022metaglyph}
Ying L, Shu X, Deng D, Yang Y, Tang T, Yu L, Wu Y. MetaGlyph: Automatic
  generation of metaphoric glyph-based visualization. \emph{{IEEE} Transactions
  on Visualization and Computer Graphics}, 2023, 29(1): 331--341.

\bibitem{guo2023edit}
Guo Y, Han Q, Lou Y, Wang Y, Liu C, Yuan X.
\newblock Edit-History Vis: An Interactive Visual Exploration and Analysis on
  Wikipedia Edit History.
\newblock In \emph{Proceedings of the IEEE Pacific Visualization Symposium},
  157--166, 2023.

\bibitem{tu2022phrasemap}
Tu Y, Qiu R, Wang YS, Yen PY, Shen HW. PhraseMap: Attention-Based Keyphrases
  Recommendation for Information Seeking. \emph{{IEEE} Transactions on
  Visualization and Computer Graphics}, 2022, \doi{10.1109/TVCG.2022.3225114},
  to be published.

\bibitem{li2021nbsearch}
Li X, Wang Y, Wang H, Wang Y, Zhao J.
\newblock {Nbsearch}: Semantic search and visual exploration of computational
  notebooks.
\newblock In \emph{Proceedings of the CHI conference on human factors in
  computing systems}, 1--14, 2021.

\bibitem{narechania2021vitality}
Narechania A, Karduni A, Wesslen R, Wall E. Vitality: Promoting serendipitous
  discovery of academic literature with transformers \& visual analytics.
  \emph{{IEEE} Transactions on Visualization and Computer Graphics}, 2022,
  28(1): 486--496.

\bibitem{shi2022medchemlens}
Shi C, Nie F, Hu Y, Xu Y, Chen L, Ma X, Luo Q. MedChemLens: An Interactive
  Visual Tool to Support Direction Selection in Interdisciplinary Experimental
  Research of Medicinal Chemistry. \emph{{IEEE} Transactions on Visualization
  and Computer Graphics}, 2023, 29(1): 63--73.

\bibitem{resck2022legalvis}
Resck LE, Ponciano JR, Nonato LG, Poco J. Legalvis: Exploring and inferring
  precedent citations in legal documents. \emph{{IEEE} Transactions on
  Visualization and Computer Graphics}, 2023, 29(6): 3105--3120.

\bibitem{zhang2020teddy}
Zhang X, Engel J, Evensen S, Li Y, Demiralp {\c{C}}, Tan WC.
\newblock Teddy: A system for interactive review analysis.
\newblock In \emph{Proceedings of the CHI conference on human factors in
  computing systems}, 1--13, 2020.

\bibitem{wu2023liveretro}
Wu Y, Xu Y, Gao S, Wang X, Song W, Nie Z, Fan X, Li Q. LiveRetro: Visual
  Analytics for Strategic Retrospect in Livestream E-Commerce. \emph{arXiv
  preprint arXiv:2307.12213}, 2023.

\bibitem{ouyang2023leveraging}
Ouyang Y, Wu Y, Wang H, Zhang C, Cheng F, Jiang C, Jin L, Cao Y, Li Q.
  Leveraging Historical Medical Records as a Proxy via Multimodal Modeling and
  Visualization to Enrich Medical Diagnostic Learning. \emph{arXiv preprint
  arXiv:2307.12199}, 2023.

\bibitem{tu2023sdrquerier}
Tu Y, Li O, Wang J, Shen HW, Powa{\l}ko P, Tomescu-Dubrow I, Slomczynski KM,
  Blanas S, Jenkins JC. {SDRQuerier}: A Visual Querying Framework for
  Cross-National Survey Data Recycling. \emph{{IEEE} Transactions on
  Visualization and Computer Graphics}, 2023, 29(6): 2862--2874.

\bibitem{chen2023iball}
Chen Z, Yang Q, Shan J, Lin T, Beyer J, Xia H, Pfister H.
\newblock iBall: Augmenting Basketball Videos with Gaze-moderated Embedded
  Visualizations.
\newblock In \emph{Proceedings of the CHI conference on human factors in
  computing systems}, 1--18, 2023.

\bibitem{chen2023sporthesia}
Chen Z, Yang Q, Xie X, Beyer J, Xia H, Wu Y, Pfister H. Sporthesia: Augmenting
  sports videos using natural language. \emph{{IEEE} Transactions on
  Visualization and Computer Graphics}, 2023, 29(1): 918--928.

\bibitem{tu2021keywordmap}
Tu Y, Xu J, Shen HW.
\newblock {KeywordMap}: Attention-based visual exploration for keyword
  analysis.
\newblock In \emph{Proceedings of the IEEE Pacific Visualization Symposium},
  206--215, 2021.

\bibitem{liu2021advisor}
Liu C, Han Y, Jiang R, Yuan X.
\newblock Advisor: Automatic visualization answer for natural-language question
  on tabular data.
\newblock In \emph{Proceedings of the IEEE Pacific Visualization Symposium},
  11--20, 2021.

\bibitem{shen2023data}
Shen L, Zhang Y, Zhang H, Wang Y. Data player: Automatic generation of data
  videos with narration-animation interplay. \emph{{IEEE} Transactions on
  Visualization and Computer Graphics}, 2024, 30(1): 1--11, accepted by VIS
  2023.

\bibitem{xiao2023let}
Xiao S, Huang S, Lin Y, Ye Y, Zeng W. Let the chart spark: Embedding semantic
  context into chart with text-to-image generative model. \emph{{IEEE}
  Transactions on Visualization and Computer Graphics}, 2024, 30(1): 1--11,
  accepted by VIS 2023.

\bibitem{singh2020stl}
Singh H, Shekhar S.
\newblock {STL-CQA}: Structure-based transformers with localization and
  encoding for chart question answering.
\newblock In \emph{Proceedings of the Conference on Empirical Methods in
  Natural Language Processing}, 3275--3284, 2020.

\bibitem{ma2021towards}
Ma W, Zhang H, Yan S, Yao G, Huang Y, Li H, Wu Y, Jin L.
\newblock Towards an efficient framework for data extraction from chart images.
\newblock In \emph{International Conference on Document Analysis and
  Recognition}, 583--597, 2021.

\bibitem{song2023vividgraph}
Song S, Li C, Sun Y, Wang C. Vividgraph: Learning to extract and redesign
  network graphs from visualization images. \emph{{IEEE} Transactions on
  Visualization and Computer Graphics}, 2023, 29(7): 3169--3181.

\bibitem{chen2019towards}
Chen Z, Wang Y, Wang Q, Wang Y, Qu H. Towards automated infographic design:
  Deep learning-based auto-extraction of extensible timeline. \emph{{IEEE}
  Transactions on Visualization and Computer Graphics}, 2020, 26(1): 917--926.

\bibitem{sultanum2023datatales}
Sultanum N, Srinivasan A. DataTales: Investigating the use of Large Language
  Models for Authoring Data-Driven Articles. \emph{{IEEE} Transactions on
  Visualization and Computer Graphics}, 2024, 30(1): 1--11, accepted by VIS
  2023.

\bibitem{liu2023autotitle}
Liu C, Guo Y, Yuan X. AutoTitle: An Interactive Title Generator for
  Visualizations. \emph{{IEEE} Transactions on Visualization and Computer
  Graphics}, 2023, \doi{10.1109/TVCG.2023.3290241}, to be published.

\bibitem{song2023gvqa}
Song S, Chen J, Li C, Wang C.
\newblock {GVQA}: Learning to Answer Questions about Graphs with Visualizations
  via Knowledge Base.
\newblock In \emph{Proceedings of the CHI conference on human factors in
  computing systems}, 1--16, 2023.

\bibitem{adhikary2021text}
Adhikary J, Vertanen K. Text entry in virtual environments using speech and a
  midair keyboard. \emph{{IEEE} Transactions on Visualization and Computer
  Graphics}, 2021, 27(5): 2648--2658.

\bibitem{card1999readings}
Card SK, Mackinlay J, Shneiderman B.
\newblock \emph{Readings in information visualization: using vision to think}.
\newblock Morgan Kaufmann, 1999.

\bibitem{zhou2023comprehensive}
Zhou C, Li Q, Li C, Yu J, Liu Y, Wang G, Zhang K, Ji C, Yan Q, He L, Peng H, Li
  J, Wu J, Liu Z, Xie P, Xiong C, Pei J, Yu PS, Sun L. A comprehensive survey
  on pretrained foundation models: A history from bert to chatgpt. \emph{arXiv
  preprint arXiv:2302.09419}, 2023.

\bibitem{chen2019lassonet}
Chen Z, Zeng W, Yang Z, Yu L, Fu CW, Qu H. LassoNet: Deep lasso-selection of 3D
  point clouds. \emph{{IEEE} Transactions on Visualization and Computer
  Graphics}, 2020, 26(1): 195--204.

\bibitem{ottley2019follow}
Ottley A, Garnett R, Wan R. Follow the clicks: Learning and anticipating mouse
  interactions during exploratory data analysis. \emph{Computer Graphics
  Forum}, 2019, 38(3): 41--52.

\bibitem{brown2014finding}
Brown ET, Ottley A, Zhao H, Lin Q, Souvenir R, Endert A, Chang R. Finding
  waldo: Learning about users from their interactions. \emph{{IEEE}
  Transactions on Visualization and Computer Graphics}, 2014, 20(12):
  1663--1672.

\bibitem{Wexler2020WhatIf}
Wexler J, Pushkarna M, Bolukbasi T, Wattenberg M, Viegas F, Wilson J. The
  what-if tool: Interactive probing of machine learning models. \emph{{IEEE}
  Transactions on Visualization and Computer Graphics}, 2020, 26(1): 56--65.

\bibitem{houlsby2019parameter}
Houlsby N, Giurgiu A, Jastrzebski S, Morrone B, De~Laroussilhe Q, Gesmundo A,
  Attariyan M, Gelly S.
\newblock Parameter-efficient transfer learning for NLP.
\newblock In \emph{International Conference on Machine Learning}, 2790--2799,
  2019.

\bibitem{hu2021lora}
Hu EJ, Shen Y, Wallis P, Allen-Zhu Z, Li Y, Wang S, Wang L, Chen W. Lora:
  Low-rank adaptation of large language models. \emph{arXiv preprint
  arXiv:2106.09685}, 2021.

\bibitem{adapterhub}
AdapterHub. \url{https://adapterhub.ml/}, {L}ast accessed 2023-10-01.

\bibitem{wei2022chain}
Wei J, Wang X, Schuurmans D, Bosma M, Xia F, Chi E, Le QV, Zhou D, et~al..
  Chain-of-thought prompting elicits reasoning in large language models.
  \emph{Advances in Neural Information Processing Systems}, 2022, 35:
  24824--24837.

\bibitem{raffel2020T5}
Raffel C, Shazeer N, Roberts A, Lee K, Narang S, Matena M, Zhou Y, Li W, Liu
  PJ. Exploring the limits of transfer learning with a unified text-to-text
  transformer. \emph{The Journal of Machine Learning Research}, 2020, 21(1):
  5485--5551.

\bibitem{wang2022towards}
Wang Y, Hou Z, Shen L, Wu T, Wang J, Huang H, Zhang H, Zhang D. Towards natural
  language-based visualization authoring. \emph{IEEE Transactions on
  Visualization and Computer Graphics}, 2023, 29(1): 1222--1232.

\bibitem{schwartz2020green}
Schwartz R, Dodge J, Smith NA, Etzioni O. Green ai. \emph{Communications of the
  ACM}, 2020, 63(12): 54--63.

\bibitem{zhou2023lima}
Zhou C, Liu P, Xu P, Iyer S, Sun J, Mao Y, Ma X, Efrat A, Yu P, Yu L, et~al..
  Lima: Less is more for alignment. \emph{arXiv preprint arXiv:2305.11206},
  2023.

\bibitem{zhou2024cluster}
Zhou Y, Yang W, Chen J, Chen C, Shen Z, Luo X, Yu L, Liu S. Cluster-Aware Grid
  Layout. \emph{{IEEE} Transactions on Visualization and Computer Graphics},
  2024, 30(1): 1--11, accepted by VIS 2023.

\bibitem{Yang2020Interactive}
Yang W, Wang X, Lu J, Dou W, Liu S. Interactive Steering of Hierarchical
  Clustering. \emph{{IEEE} Transactions on Visualization and Computer
  Graphics}, 2020, to be published,
  \href{http://dx.doi.org/10.1109/TVCG.2020.2995100}{doi:
  \textcolor{blue}{10.1109/TVCG.2020.2995100}}.

\bibitem{abadi2016tensorflow}
Abadi M, Agarwal A, Barham P, Brevdo E, Chen Z, Citro C, Corrado GS, Davis A,
  Dean J, Devin M, et~al.. Tensorflow: Large-scale machine learning on
  heterogeneous distributed systems. \emph{CoRR}, 2016, abs/1603.04467,
  \doi{10.48550/arXiv.1603.04467}.

\bibitem{ma2009situ}
Ma KL. In situ visualization at extreme scale: Challenges and opportunities.
  \emph{IEEE Computer Graphics and Applications}, 2009, 29(6): 14--19.

\bibitem{rapp2022image}
Rapp T, Peters C, Dachsbacher C. Image-based visualization of large volumetric
  data using moments. \emph{{IEEE} Transactions on Visualization and Computer
  Graphics}, 2022, 28(6): 2314--2325.

\bibitem{richer2022scalability}
Richer G, Pister A, Abdelaal M, Fekete JD, Sedlmair M, Weiskopf D. Scalability
  in visualization. \emph{{IEEE} Transactions on Visualization and Computer
  Graphics}, 2022, \doi{10.1109/TVCG.2022.3231230}, to be published.

\bibitem{dong2022survey}
Dong Q, Li L, Dai D, Zheng C, Wu Z, Chang B, Sun X, Xu J, Sui Z. A survey for
  in-context learning. \emph{arXiv preprint arXiv:2301.00234}, 2022.

\bibitem{Liu2018Visual}
Liu S, Xiao J, Liu J, Wang X, Wu J, Zhu J. Visual diagnosis of tree boosting
  methods. \emph{{IEEE} Transactions on Visualization and Computer Graphics},
  2018, 24(1): 163--173.

\bibitem{yuan2022visual}
Yuan J, Liu M, Tian F, Liu S. Visual Analysis of Neural Architecture Spaces for
  Summarizing Design Principles. \emph{{IEEE} Transactions on Visualization and
  Computer Graphics}, 2023, 29(1): 288--298.

\bibitem{khayat2019vassl}
Khayat M, Karimzadeh M, Zhao J, Ebert DS. {VASSL}: A Visual Analytics Toolkit
  for Social Spambot Labeling. \emph{{IEEE} Transactions on Visualization and
  Computer Graphics}, 2020, 26(1): 874--883.

\bibitem{bernard2018towards}
Bernard J, Zeppelzauer M, Lehmann M, M{\"u}ller M, Sedlmair M. Towards
  User-Centered Active Learning Algorithms. \emph{Computer Graphics Forum},
  2019, 37(3): 121--132.

\bibitem{yang2022diagnosing}
Yang W, Ye X, Zhang X, Xiao L, Xia J, Wang Z, Zhu J, Pfister H, Liu S.
  Diagnosing Ensemble Few-Shot Classifiers. \emph{{IEEE} Transactions on
  Visualization and Computer Graphics}, 2022, 28(9): 3292--3306,
  \doi{10.1109/TVCG.2022.3182488}.

\bibitem{zhou2022learnware}
Zhou ZH, Tan ZH. Learnware: Small models do big. \emph{arXiv preprint
  arXiv:2210.03647}, 2022.

\bibitem{huggingface}
HuggingFace. \url{https://huggingface.co/models}, {L}ast accessed 2023-10-01.

\bibitem{Wang2020Visual}
Wang Q, Yuan J, Chen S, Su H, Qu H, Liu S. Visual Genealogy of Deep Neural
  Networks. \emph{{IEEE} Transactions on Visualization and Computer Graphics},
  2020, to be published,
  \href{http://dx.doi.org/10.1109/TVCG.2019.2921323}{doi:
  \textcolor{blue}{10.1109/TVCG.2019.2921323}}.

\bibitem{Cao2020Analyzing}
Cao K, Liu M, Su H, Wu J, Zhu J, Liu S. Analyzing the Noise Robustness of Deep
  Neural Networks. \emph{{IEEE} Transactions on Visualization and Computer
  Graphics}, 2020, to be published,
  \href{http://dx.doi.org/10.1109/TVCG.2020.2969185}{doi:
  \textcolor{blue}{10.1109/TVCG.2020.2969185}}.

\bibitem{liu2018robustness}
Liu M, Liu S, Su H, Cao K, Zhu J.
\newblock Analyzing the noise robustness of deep neural networks.
\newblock In \emph{IEEE Conference on Visual Analytics Science and Technology
  (VAST)}, 60--71. IEEE, 2018.

\bibitem{qiu2022docflow}
Qiu R, Tu Y, Wang YS, Yen PY, Shen HW. DocFlow: A Visual Analytics System for
  Question-based Document Retrieval and Categorization. \emph{{IEEE}
  Transactions on Visualization and Computer Graphics}, 2022,
  \doi{10.1109/TVCG.2022.3219762}, to be published.

\bibitem{blei2003latent}
Blei DM, Ng AY, Jordan MI. Latent dirichlet allocation. \emph{Journal of
  Machine Learning Research}, 2003, 3(Jan): 993--1022.

\bibitem{lowe1999object}
Lowe DG.
\newblock Object recognition from local scale-invariant features.
\newblock In \emph{Proceedings of the IEEE International Conference on Computer
  Vision}, volume~2, 1150--1157, 1999.

\bibitem{roziere2023code}
Rozi{\`e}re B, Gehring J, Gloeckle F, Sootla S, Gat I, Tan XE, Adi Y, Liu J,
  Remez T, Rapin J, et~al.. Code llama: Open foundation models for code.
  \emph{arXiv preprint arXiv:2308.12950}, 2023.

\bibitem{austin2021program}
Austin J, Odena A, Nye M, Bosma M, Michalewski H, Dohan D, Jiang E, Cai C,
  Terry M, Le Q, et~al.. Program synthesis with large language models.
  \emph{arXiv preprint arXiv:2108.07732}, 2021.

\bibitem{bostock2011d3}
Bostock M, Ogievetsky V, Heer J. D$^3$ data-driven documents. \emph{{IEEE}
  Transactions on Visualization and Computer Graphics}, 2011, 17(12):
  2301--2309.

\bibitem{hunter2007matplotlib}
Hunter JD. Matplotlib: A 2D graphics environment. \emph{Computing in Science \&
  Engineering}, 2007, 9(03): 90--95.

\bibitem{kwon2019deep}
Kwon OH, Ma KL. A deep generative model for graph layout. \emph{{IEEE}
  Transactions on Visualization and Computer Graphics}, 2020, 26(1): 665--675.

\bibitem{zamfirescu2023johnny}
Zamfirescu-Pereira J, Wong RY, Hartmann B, Yang Q.
\newblock Why Johnny can’t prompt: how non-AI experts try (and fail) to
  design LLM prompts.
\newblock In \emph{Proceedings of the CHI conference on human factors in
  computing systems}, 1--21, 2023.

\bibitem{pryzant2023automatic}
Pryzant R, Iter D, Li J, Lee YT, Zhu C, Zeng M. Automatic prompt optimization
  with "gradient descent" and beam search. \emph{arXiv preprint
  arXiv:2305.03495}, 2023.

\bibitem{jing2019neural}
Jing Y, Yang Y, Feng Z, Ye J, Yu Y, Song M. Neural style transfer: A review.
  \emph{IEEE transactions on visualization and computer graphics}, 2019,
  26(11): 3365--3385.

\bibitem{abdal2019image2stylegan}
Abdal R, Qin Y, Wonka P.
\newblock {Image2StyleGAN}: How to Embed Images Into the StyleGAN Latent Space?
\newblock In \emph{Proceedings of the IEEE/CVF International Conference on
  Computer Vision}, 4432--4441, 2019.

\bibitem{antol2015vqa}
Antol S, Agrawal A, Lu J, Mitchell M, Batra D, Zitnick CL, Parikh D.
\newblock {VQA}: Visual Question Answering.
\newblock In \emph{Proceedings of the IEEE International Conference on Computer
  Vision}, 2425--2433, 2015.

\bibitem{anil2023palm}
Anil R, Dai AM, Firat O, Johnson M, Lepikhin D, Passos A, Shakeri S, Taropa E,
  Bailey P, Chen Z, et~al.. Palm 2 technical report. \emph{arXiv preprint
  arXiv:2305.10403}, 2023.

\bibitem{zhao2021preserving}
{Zhao} Y, {Jiang} H, {Chen} Q, {Qin} Y, {Xie} H, {Wu} Y, {Liu} S, {Zhou} Z,
  {Xia} J, {Zhou} F. Preserving Minority Structures in Graph Sampling.
  \emph{IEEE Transactions on Visualization and Computer Graphics}, 2021, 27(2):
  1698--1708.

\bibitem{yuan2020evaluation}
{Yuan} J, {Xiang} S, {Xia} J, {Yu} L, {Liu} S. Evaluation of Sampling Methods
  for Scatterplots. \emph{IEEE Transactions on Visualization and Computer
  Graphics}, 2021, 27(2): 1720--1730.

\bibitem{pan2023drag}
Pan X, Tewari A, Leimk{\"u}hler T, Liu L, Meka A, Theobalt C.
\newblock Drag Your GAN: Interactive Point-Based Manipulation on the Generative
  Image Manifold.
\newblock In \emph{ACM SIGGRAPH Conference Proceedings}, 1--11, 2023.

\bibitem{wang2023survey}
Wang L, Ma C, Feng X, Zhang Z, Yang H, Zhang J, Chen Z, Tang J, Chen X, Lin Y,
  et~al.. A survey on large language model based autonomous agents. \emph{arXiv
  preprint arXiv:2308.11432}, 2023.

\bibitem{park2023generative}
Park JS, O'Brien JC, Cai CJ, Morris MR, Liang P, Bernstein MS. Generative
  agents: Interactive simulacra of human behavior. \emph{arXiv preprint
  arXiv:2304.03442}, 2023.

\end{thebibliography}

\subsection*{Author biography}

\begin{biography}[portraits/weikai.jpg]{Weikai Yang} {is a Ph.D. candidate at Tsinghua University. His research interests include visual text analytics and interactive machine learning. He received a B.S. degree from Tsinghua University.
yangwk21@mails.tsinghua.edu.cn
}
\end{biography}

\vspace{1cm}

\begin{biography}[portraits/mengchen.jpg]{Mengchen Liu} {is a Senior Researcher at Microsoft. His research interests include explainable AI and computer vision. He received a B.S. in Electronics Engineering and a Ph.D. in Computer Science from Tsinghua University.
He has served as a PC member and reviewer for various conferences and journals.
mengcliu@microsoft.com
}
\end{biography}

\vspace{1cm}

\begin{biography}[portraits/wangzheng.jpg]{Zheng Wang} {is currently working toward a graduate degree at Tsinghua University.
zheng-wa19@mails.tsinghua.edu.cn 
}
\end{biography}

\vspace{1cm}

\begin{biography}[portraits/shixia.jpg]{Shixia Liu} {is a professor at Tsinghua University. Her research interests include visual text analytics, visual social analytics, interactive machine learning, and text mining. She worked as a research staff member at IBM China Research Lab and a lead researcher at Microsoft Research Asia.
She received a B.S. and M.S. from Harbin Institute of Technology, a Ph.D. from Tsinghua University.
She is a fellow of IEEE and an associate editor-in-chief of IEEE Trans. Vis. Comput. Graph.
shixia@tsinghua.edu.cn
}
\end{biography}
\vspace{2cm}
\vspace*{2.6em}
\subsection*{Graphical abstract}

\begin{figure}[!htb]
\centering
\includegraphics[width=8cm]{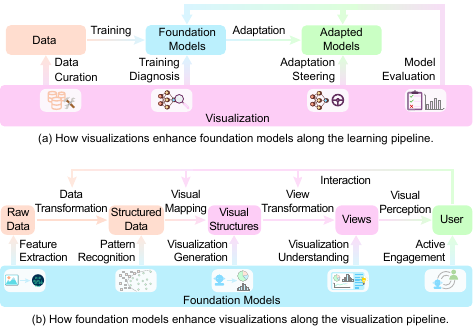}
\label{fig:abstract}
\end{figure}

\end{document}